%% file: neurips_2026.tex
\PassOptionsToPackage{numbers,compress}{natbib}
\documentclass{article}

\usepackage[preprint]{neurips_2026}

\usepackage[utf8]{inputenc}
\usepackage[T1]{fontenc}
\usepackage{hyperref}
\usepackage{url}
\usepackage{microtype}
\usepackage{graphicx}
\usepackage{subcaption}
\usepackage{booktabs}
\usepackage{enumitem}
\usepackage[table]{xcolor}
\usepackage{multirow}
\usepackage{makecell}
\usepackage{colortbl}
\usepackage{tabularx}
\usepackage{amsfonts}
\usepackage{amsmath}
\usepackage{amssymb}
\usepackage{mathtools}
\usepackage{amsthm}
\usepackage{nicefrac}
\usepackage[capitalize,noabbrev]{cleveref}
\usepackage{tcolorbox}
\usepackage{algorithm}
\usepackage{algpseudocode}

\newcolumntype{Y}{>{\centering\arraybackslash}X}
\newcolumntype{B}{>{\centering\arraybackslash}m{6.8em}}
\newcolumntype{M}{>{\centering\arraybackslash}m{6.6em}}
\newcolumntype{N}{>{\centering\arraybackslash}m{4.2em}}

\theoremstyle{plain}

\theoremstyle{definition}

\theoremstyle{remark}

\newcommand{\best}[1]{\textcolor{red}{\textbf{#1}}}
\newcommand{\second}[1]{\textcolor{blue}{\textbf{#1}}}
\newcommand{\norm}[2]{#1$_{\scriptscriptstyle\pm #2}$}

\DeclareRobustCommand{\figref}[1]{\hyperref[#1]{Figure~\ref*{#1}}}
\DeclareRobustCommand{\figpartref}[2]{\hyperref[#1]{Figure~\ref*{#1}(#2)}}
\DeclareRobustCommand{\tabref}[1]{\hyperref[#1]{Table~\ref*{#1}}}
\DeclareRobustCommand{\apptabref}[1]{\hyperref[#1]{Appendix Table~\ref*{#1}}}
\DeclareRobustCommand{\appref}[1]{\hyperref[#1]{Appendix~\ref*{#1}}}
\definecolor{tableheadergray}{RGB}{224,224,230}
\definecolor{lightcyan}{rgb}{0.88, 1, 1}
\definecolor{hlfirstbg}{RGB}{203,226,255}
\definecolor{hlsecondbg}{RGB}{235,244,255}
\definecolor{myred}{rgb}{1, 0, 0}
\definecolor{myblue}{rgb}{0, 0, 1}
\newcommand{\tworowhead}[1]{\raisebox{-0.55\baselineskip}{\textbf{#1}}}
\newcommand{\hlfirst}[1]{\begingroup\setlength{\fboxsep}{1.3pt}\colorbox{hlfirstbg}{\textbf{#1}}\endgroup}
\newcommand{\hlsecond}[1]{\begingroup\setlength{\fboxsep}{1.3pt}\colorbox{hlsecondbg}{#1}\endgroup}
\newcommand{\sourcebadge}[1]{\begingroup\setlength{\fboxsep}{1.2pt}\colorbox{gray!70}{\textcolor{white}{\small\textbf{#1}}}\endgroup}
\newcommand{\baselinecell}[2]{\makecell[c]{#1\\[-0.1em]{\sourcebadge{#2}}}}
\newcommand{\onerowmethod}[1]{#1}
\newcommand{\alignmentrowshift}{\noalign{\vskip 3.3pt}}
\newcommand{\dogmarowshift}{\noalign{\vskip 1.65pt}}

\newcommand{\includegraphicssafe}[2][]{%
  \IfFileExists{#2}{%
    \includegraphics[#1]{#2}%
  }{%
    \fbox{%
      \parbox[c][0.22\textheight][c]{0.9\linewidth}{%
        \centering Missing figure file\\\texttt{#2}%
      }%
    }%
  }%
}

\title{DOGMA: Weaving Structural Information into Data-centric Single-cell Transcriptomics Analysis}

\author{\normalfont
Ru Zhang$^{1}$ \quad Xunkai Li$^{1}$ \quad Yaxin Deng$^{1}$ \quad Sicheng Liu$^{1}$ \quad Daohan Su$^{1}$\\
Qiangqiang Dai$^{1}$ \quad Hongchao Qin$^{1}$ \quad Rong-Hua Li$^{1}$ \quad Guoren Wang$^{1}$ \quad Jia Li$^{2}$\\
$^{1}$Department of Computer Science, Beijing Institute of Technology, Beijing, China\\
$^{2}$The Hong Kong University of Science and Technology (GZ), Guangzhou, China\\
\textbf{Correspondence to:} Rong-Hua Li \texttt{<lironghuabit@126.com>}
}

\begin{document}

\maketitle

\begin{abstract}
Recently, data-centric AI methodology has been a dominant paradigm in single-cell transcriptomics analysis, which treats data representation rather than model complexity as the fundamental bottleneck.
In the review of current studies, earlier sequence methods treat cells as independent entities and adapt prevalent ML models to analyze their directly inherited sequence data.
Despite their simplicity and intuition, these methods overlook the latent intercellular relationships driven by the functional mechanisms of biological systems and the inherent quality issues of the raw sequencing data.
Therefore, a series of structured methods has emerged.
Although they employ various heuristic rules to capture intricate intercellular relationships and enhance the raw sequencing data, these methods often neglect biological prior knowledge.
This omission incurs substantial overhead and yields suboptimal graph representations, hindering the utility of ML models.

To address these issues, we propose DOGMA, a data-centric framework designed for the structural reshaping and semantic enhancement of raw data through multi-level biological prior knowledge. Transcending reliance on purely data-driven heuristics, DOGMA provides a prior-guided graph construction pipeline that integrates statistical alignment with Cell Ontology and phylogenetic structure for biologically grounded cell-graph construction and robust cross-species alignment. Furthermore, Gene Ontology is utilized to bridge the feature-level semantic gap by incorporating functional priors. In complex multi-species and multi-organ benchmarks, DOGMA exhibits strong robustness in strict zero-shot cell-type evaluation and sample efficiency while using substantially lower GPU memory and inference time in downstream evaluation.
\end{abstract}

\section{Introduction}

Elucidating cellular functional characteristics and their complex collaborative mechanisms lies at the heart of understanding biological systems \cite{regev2017science}. Breakthroughs in single-cell RNA sequencing (scRNA-seq) technologies have revolutionized this exploration by enabling the quantification of genome-wide expression at single-cell resolution \cite{macosko2015highly, zheng2017massively}. In response to these burgeoning analytical demands, the field of single-cell analysis is undergoing a profound paradigm shift: moving from a Model-Centric approach, which blindly pursues architectural complexity, to a Data-Centric paradigm that prioritizes data quality and structural representation \cite{zha2023datacentric}.

This paradigm shift reveals a key insight: as deep learning architectures mature, the performance bottleneck in representation learning often stems not from insufficient model capacity, but from the quality and structural integrity of the input data \cite{zha2023datacentric}. For raw single-cell sequencing data, this bottleneck manifests in two aspects. First, the raw data suffers from significant intrinsic imperfections, including high dimensionality, extreme sparsity, and inevitable technical noise (e.g., dropout events) \cite{kharchenko2014bayesian, luecken2019current}. Second, cells are not isolated entities but reside within complex biological networks. While inherent biological functional correlations exist between cells, raw data often lacks an explicit representation of these associations.

Failure to fundamentally address these issues means that merely expanding neural network depth will not only encounter diminishing marginal returns but also risk overfitting statistical artifacts, leading the model to learn spurious biological correlations. Therefore, constructing a unified Data-Centric framework capable of simultaneously achieving data denoising and structural reshaping has become an urgent imperative. However, existing mainstream methods, whether structure-agnostic sequence-based modeling \cite{scbert, geneformer, scgpt} or current noise-constrained graph construction approaches \cite{scmognn, cellvgae, scpriorgraph}, have failed to effectively address this dual challenge.

\begin{figure*}[t]
\centering
\includegraphics[width=\textwidth]{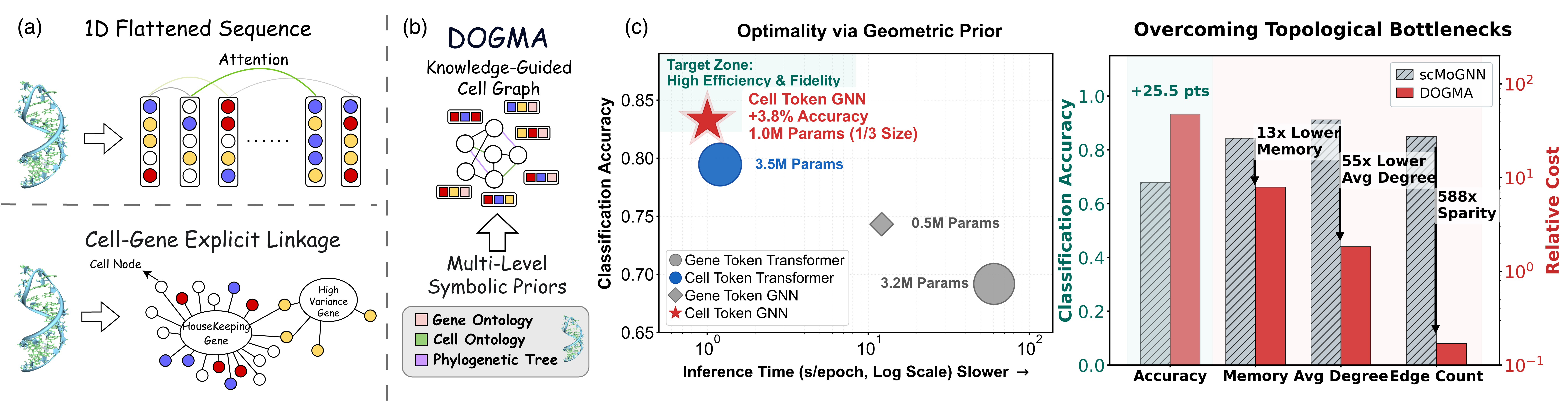}
\caption{\textbf{DOGMA: A Data-Centric Paradigm Shift.} \textbf{(a)} Data structure of current input paradigms. Sequence-based inputs lose topological structure, while heterogeneous graphs suffer from structural redundancy. \textbf{(b)} DOGMA acts as a prior-guided graph construction pipeline. It injects multi-level knowledge to produce a knowledge-guided cell graph. \textbf{(c)} Empirical validation. GNN achieves the target zone with significantly fewer parameters than Transformers, and DOGMA reduces downstream GPU-memory usage compared to scMoGNN while maintaining competitive accuracy.}
\label{fig:teaser}
\end{figure*}

Sequence-based architectures, including Transformer-style transcriptome encoders \cite{scbert, geneformer, scgpt}, treat gene expression profiles as tokenized inputs, attempting to capture latent biological patterns through large-scale pre-training. As illustrated in \figpartref{fig:teaser}{a}, however, this representation-centric view does not explicitly encode cell-cell relationships in the input, leaving downstream models to infer relational structure implicitly from sparse and noisy expression profiles \cite{luecken2019current,battaglia2018relational}.

This strategy of fitting complex models directly to noisy expression data helps explain why large-scale models, despite immense computational cost, do not always outperform simpler generative baselines such as scVI \cite{scvi} in zero-shot tasks \cite{kedzierska2025zeroshot}. Consistent with \figpartref{fig:teaser}{c}, our empirical comparison shows that scaling model size alone cannot compensate for the absence of explicit structural priors, and that such priors can yield stronger accuracy with fewer parameters than a Cell Token Transformer.

Recent ontology-aware transcriptome foundation models such as scCello~\cite{yuan2024sccello} have shown that Cell Ontology can provide valuable supervision for learning biologically meaningful cell embeddings. However, such methods mainly use ontology as a pre-training or representation-level constraint, rather than converting biological priors into an explicit, reusable cell-cell topology. This motivates a data-structure-level question: can multi-level biological priors be directly transformed into a robust graph topology for downstream learning?

Graph-based structured approaches attempt to answer this question by incorporating relational structures, but the graphs they construct still suffer from fundamental defects in two core dimensions: topological connectivity and node features, thereby limiting their effectiveness.

First, at the level of topological structure, existing graph construction strategies often rely on local statistical similarity or fragmented priors, making it difficult to impose systematic biological topological constraints. In heterogeneous graphs (e.g., scMoGNN \cite{scmognn}), introduced housekeeping gene nodes evolve into super-hub nodes, artificially bridging distinct cells and smoothing out specific differences \cite{li2018deeper}. As quantified in \figpartref{fig:teaser}{b}, these high-degree hubs precipitate an explosive growth in edge density and memory consumption, imposing severe computational bottlenecks without yielding proportional performance gains.

Meanwhile, metric-based graph construction methods (e.g., k-NN \cite{scanpy2018, cellvgae}) can be misled by batch effects, producing spurious neighborhood relationships \cite{mnn2018}. Existing prior-informed graph methods often focus on molecular-level associations \cite{scpriorgraph, scbignn}, but lack a hierarchical cell-type reference such as Cell Ontology for globally constraining cell-cell topology \cite{cellontology}. At the feature level, most pipelines still rely on HVG- or PCA-processed numerical representations \cite{luecken2019current, seurat2019}, which provide limited functional semantics. Without external knowledge bases such as Gene Ontology \cite{geneontology} to provide explicit biological definitions, downstream models may remain more vulnerable to noise and less able to focus on biologically meaningful signals.

To bridge this semantic gap, we introduce \textbf{DOGMA} (\textbf{D}ata-centric \textbf{O}ntology-\textbf{G}uided \textbf{M}odeling \textbf{A}pproach). This data-centric framework reformulates graph construction from unverified statistical inference to prior-guided cell-graph construction. Unlike methods relying solely on heuristics, DOGMA injects multi-level symbolic priors to regularize the graph construction process.

Specifically, we construct a composite cell topology jointly constrained by statistical alignment and multilayer prior knowledge: MNN is employed for initial \textbf{batch-invariant alignment}, the Cell Ontology is integrated to enforce \textbf{biological relatedness}, and phylogenetic trees are incorporated to capture \textbf{cross-species lineage conservation}. Furthermore, we leverage the Gene Ontology for feature-level data enhancement.

Empirically, we validate the effectiveness of DOGMA through extensive experiments. By aligning cells via universal biological knowledge, DOGMA establishes a robust and scalable structural foundation for the next generation of single-cell analysis.

Our main contributions are summarized as follows.

\begin{itemize}[leftmargin=1.5em]
\item \textbf{New Perspective.} We formulate single-cell graph construction as a data-centric problem rather than a model-scaling problem. DOGMA replaces purely metric-based neighborhood heuristics with prior-guided, ontology-aware structure construction, showing that a knowledge-anchored cell topology can reduce the impact of noisy and sparse sequencing measurements.
\item \textbf{New Prior-Guided Pipeline.} We develop a scalable pipeline that reshapes raw scRNA-seq data into a biologically constrained cell graph. The pipeline combines MNN statistical anchors, Cell Ontology or HCAO cell-type semantics, phylogenetic cross-species constraints, and Gene Ontology feature augmentation. This design turns heterogeneous biological priors into a reusable graph input for standard GNN backbones and supports evaluation across cross-species and cross-organ settings.
\item \textbf{Impressive Performance.} DOGMA achieves the best metadata-average accuracy across Brain, Human, and Multi, demonstrating consistent gains in cell metadata prediction. It also achieves the best strict zero-shot ARI on all three benchmarks, improving over the strongest competing result by 0.0142 on Brain, 0.0301 on Human, and 0.0282 on Multi. DOGMA delivers these gains with downstream inference time and reserved GPU memory that are tens to thousands of times lower than heavy representation-centric and graph-structured baselines.
\end{itemize}

\section{Preliminaries}
\label{sec:preliminary}

In this section, we formally define the single-cell data structures, the external symbolic knowledge bases, and the core research problem of prior-guided cell-graph construction.

\textbf{Notation.}
Let $\mathcal{C} = \{c_1, \dots, c_N\}$ denote a set of $N$ single cells and $G = \{g_1, \dots, g_M\}$ denote a set of $M$ genes. The raw input consists of three components. (1) A gene expression matrix $\mathbf{X}^{raw} \in \mathbb{R}^{N \times M}$, where the row vector $\mathbf{x}^{raw}_i \in \mathbb{R}^M$ represents the gene expression profile of cell $c_i$. (2) Associated metadata, where each cell $c_i$ is annotated with a species domain label $s_i \in \mathcal{S}$. (3) Partially observed cell-type annotations, where only training/reference cells may carry an optional label $y_i$. We denote the availability of a cell-type label by $m_i \in \{0,1\}$, with $m_i=1$ for labeled training/reference cells and $m_i=0$ otherwise. Based on these inputs, our fundamental objective is to infer a robust adjacency matrix $\mathbf{A} \in \{0,1\}^{N \times N}$ that captures intrinsic biological connectivity and transcends noise-induced artifacts, serving as the topological backbone for the following graph representation learning.

\textbf{Symbolic Knowledge Definition.}
To mitigate the inherent noise and sparsity in raw sequencing data, we leverage multi-level structured biological priors:
(1) \textit{Cell Ontology (CL):} We define the Cell Ontology as a Directed Acyclic Graph (DAG) $\mathcal{G}_{CL} = (\mathcal{V}_{CL}, \mathcal{E}_{CL})$, where nodes $\mathcal{V}_{CL}$ represent standardized cell types and edges $\mathcal{E}_{CL}$ represent hierarchical relationships. For reference cells with $m_i=1$, the labels serve as direct indices $y_i \in \mathcal{V}_{CL}$.
(2) \textit{Gene Ontology (GO):} We define the Gene Ontology as a DAG $\mathcal{G}_{GO} = (\mathcal{V}_{GO}, \mathcal{E}_{GO})$, where nodes represent gene functional terms. The mapping between genes and functions is provided by the GO database: for each gene feature $g_j \in G$, its associated functional terms correspond to a subset of nodes $\mathcal{V}_{g_j} \subset \mathcal{V}_{GO}$.
(3) \textit{Phylogeny:} We model cross-species evolutionary relationships via a phylogenetic tree $\mathcal{T}_{phy} = (\mathcal{V}_{phy}, \mathcal{E}_{phy})$. The species labels $s_i$ correspond to leaf nodes in this tree, where the tree distance $d_{phy}(\cdot, \cdot)$ reflects evolutionary divergence.

\textbf{Problem Formulation.}
Standard paradigms typically construct an adjacency matrix $\mathbf{A}_{\text{naive}}$ based on metric heuristics in the feature space (e.g., $k$-NN). However, due to technical noise, $\mathbf{A}_{\text{naive}}$ often fails to reflect the true biological topology.
Our goal is to replace purely heuristic graph construction with prior-guided cell-graph construction, so that the resulting cell graph remains both statistically informative and biologically plausible.

\section{Related Work}
\textbf{Deep Paradigms.}
The field has transitioned from statistical approaches to deep architectures. Representation-centric transcriptome encoders such as scBERT~\cite{scbert}, Geneformer~\cite{geneformer}, and scGPT~\cite{scgpt} adapt Transformer architectures to gene expression profiles, treating them as tokenized inputs for large-scale representation learning. While scalable, this tokenization requires massive pre-training to learn purely statistical patterns, leading to high data inefficiency without biological grounding. Conversely, Graph Neural Networks (e.g., scGNN~\cite{scgnn}) offer biologically plausible inductive biases by modeling cellular interactions, yet their efficacy is strictly bounded by the quality of the input graph structure.

\textbf{Evolution of Graph Construction Strategies.}
Current graph construction paradigms largely rely on data-driven heuristics, lacking external verification.
(1) \textit{Metric-based Heuristics.} Dominant approaches like SPRING~\cite{spring} and scGAC~\cite{scgac} construct $k$-NN graphs based on Euclidean or correlation metrics. These methods rely on the assumption that geometric proximity equals biological relatedness, rendering them vulnerable to stochastic noise and technical batch effects~\cite{luecken2019current}. Moreover, lacking a shared semantic coordinate system, these methods are typically limited to \textit{transductive} settings, failing to generalize to unseen batches or cell types.
(2) \textit{Latent-Graph Learning.} Methods like CellVGAE~\cite{cellvgae} and scBiGNN~\cite{scbignn} infer graph structures within learned latent spaces. While they improve upon raw features, the inferred topology often overfits to the internal statistical consistency of the training data rather than true biological fidelity, limiting cross-dataset transferability~\cite{saturn}.
(3) \textit{Heterogeneous Modeling.} Approaches like scMoGNN~\cite{scmognn} explicitly link cells and genes. However, high-degree gene nodes act as super-hub nodes, leading to severe information over-smoothing~\cite{li2018deeper} and prohibitive memory overheads.

\textbf{Knowledge-Informed Representation Learning.}
A growing body of work seeks to integrate biological priors.
\textit{Statistical Alignment.} Methods like scGCN~\cite{scgcn} utilize Mutual Nearest Neighbors (MNN)~\cite{mnn2018} to align distributions. However, MNN is a statistical heuristic reliant on local geometry, lacking semantic verification.
\textit{Architectural Constraints.} Recent approaches like expiMap~\cite{expimap} directly embed Gene Ontology into the neural network architecture. These methods represent a \textit{model-centric} approach, where knowledge is hard-coded as architectural constraints for interpretability.

\textbf{Our Distinction.} In contrast to these \textit{model-centric} architectures, DOGMA adopts a \textit{data-centric} paradigm. We focus on engineering the input topology itself rather than the model architecture. By rigorously combining statistical anchors with symbolic priors (Cell Ontology \& Phylogeny), we construct a universal, biologically constrained graph structure. This allows DOGMA to be used as a universal plug-and-play graph module for any standard GNN backbone, distinguishing our contribution from specialized end-to-end architectures.

\section{Methods}
\label{sec:methods}

\begin{figure*}[t!]
    \centering
    \includegraphics[width=\textwidth]{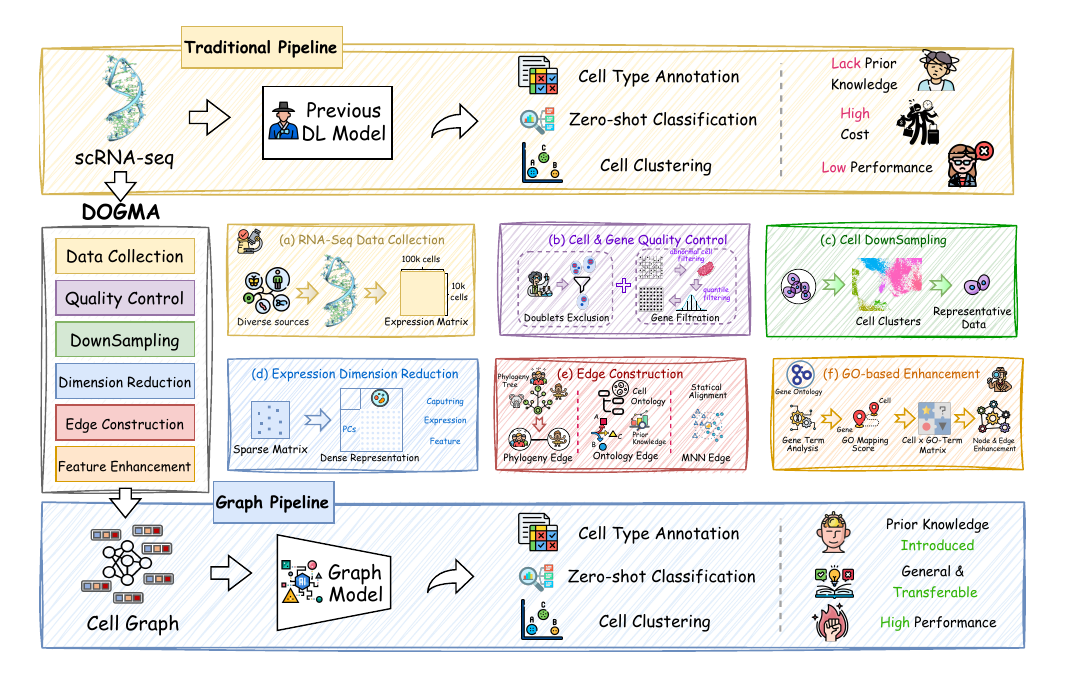}
    \caption{\textbf{The DOGMA Framework}
\textbf{(Top)} Traditional pipelines rely on black-box models that lack prior knowledge, leading to high computational costs and suboptimal performance.
\textbf{(Middle)} Our proposed data-centric workflow transforms raw scRNA-seq data into a knowledge-guided cell graph through six stages: (a--b) rigorous data curation and quality control; (c--d) representative downsampling and dimensionality reduction; (e) prior-guided topology construction integrating Phylogeny, Cell Ontology, and MNN edges; and (f) feature enhancement via Gene Ontology (GO).
\textbf{(Bottom)} The resulting Cell Graph serves as a universal, interpretable input for Graph Models, enabling high-performance analysis across strict zero-shot cell-type evaluation and clustering tasks.}
    \label{fig:framework}
\end{figure*}

DOGMA converts raw single-cell expression profiles into a biologically constrained cell graph. Given raw expression matrix $\mathbf{X}^{raw}$, species labels $\{s_i\}$, and optional cell-type labels $\{y_i\}$ for training/reference cells only, DOGMA outputs a binary adjacency matrix $\mathbf{A}$ and node representation $\mathbf{H}$ for downstream GNN training. The framework has three stages: statistical feature initialization, prior-guided topology construction, and Gene Ontology (GO)-based semantic feature fusion. In strict zero-shot evaluation, we use a label-strict transductive setting: unseen-class nodes may remain in the graph for message passing, but their cell-type labels are excluded from graph construction, supervision, classifier outputs, and prototype computation. Full mathematical and implementation details are provided in \appref{app:method_details}.
As summarized in \figref{fig:framework}, the pipeline first curates and normalizes raw scRNA-seq inputs, then constructs a prior-guided cell graph, and finally enriches node features with Gene Ontology semantics.

\subsection{Input Representation}
\label{subsec:data_pipeline}
We apply quality control, stratified downsampling, and log-normalization to CELLxGENE-derived expression matrices. Following standard single-cell pipelines~\cite{scanpy2018, seurat2019}, the normalized expression matrix is projected into a 50-dimensional PCA space $\mathbf{X}^{pca}$, which serves as the statistical view for graph construction and node representation.

\subsection{Prior-Guided Topology Construction}
\label{subsec:topology}

DOGMA constructs the cell graph through three complementary edge branches: statistical alignment, ontology-guided semantic masking, and phylogenetically stratified cross-species connection. These branches are generated independently and then merged into a single homogeneous graph for downstream message passing.

\paragraph{Statistical alignment.}
We first preserve local expression-space structure by constructing mutual-nearest-neighbor edges within each species. For species $s$, let $\mathcal{I}_s = \{i \mid s_i=s\}$ denote the index set of cells from species $s$. The alignment branch is
\begin{equation}
\mathcal{E}_{\text{Align}} =
\bigcup_{s \in \mathcal{S}}
\left\{
(i,j)
\;\middle|\;
i,j \in \mathcal{I}_s,\
j \in \mathrm{NN}^{\cos}_{k_{\text{mnn}}}(i;\mathbf{X}^{pca}),\
i \in \mathrm{NN}^{\cos}_{k_{\text{mnn}}}(j;\mathbf{X}^{pca})
\right\}.
\end{equation}
This branch supplies the statistical backbone of the graph before biological priors are injected.

\paragraph{Ontology-guided masking.}
The second branch uses cell-type ontology to restrict graph neighbors to biologically admissible candidates, but only among cells whose labels are available during graph construction. Let $s^{pca}_{ij}=\cos(\mathbf{x}^{pca}_i,\mathbf{x}^{pca}_j)$ denote PCA-space similarity and $m_i \in \{0,1\}$ indicate whether cell $i$ is a training/reference cell with an available cell-type label. Validation, test, query, or otherwise unlabeled cells have $m_i=0$. Their cell-type labels are never used to construct ontology edges. For each labeled training/reference cell, DOGMA first forms an ontology-admissible candidate set among other labeled training/reference cells and then keeps the most similar candidates:
\begin{equation}
\resizebox{0.985\linewidth}{!}{$\displaystyle
\mathcal{E}_{\text{Onto}} =
\left\{
(i,j)
\;\middle|\;
j \in \operatorname{TopK}_{c \in \mathcal{C}^{\text{Onto}}_i}\!\left(s^{pca}_{ic},\,k_{\text{onto}}\right),\!
\mathcal{C}^{\text{Onto}}_i =\!
\left\{
c \in \mathcal{C}
\;\middle|\;
m_i = m_c = 1,\ d_{O_b}(y_i,y_c) \le \tau_b
\right\}
\right\}.
$}
\end{equation}
For cross-species benchmarks (Brain and Multi), we use Cell Ontology (CL) as the semantic reference $O_b$. For the single-species multi-organ Human benchmark, we use HCAO to better capture organ-specific cell-type granularity. The ontology distance threshold $\tau_b$ is selected per benchmark from $\{1,2\}$ and capped at $\tau_b \le 2$, because larger distances may connect biologically divergent cell types as neighbors, such as endothelial cells and erythroid lineage cells, thereby introducing noisy edges into the graph topology.
Thus, ontology-derived edges are created only between training/reference nodes. Unlabeled nodes are still present in the graph, but they are connected through label-free branches such as statistical alignment and, when applicable, cross-species bridging.

\paragraph{Cross-species connection.}
For cross-species benchmarks, DOGMA builds $\mathcal{E}_{\text{Phy}}$ by projecting each species pair into a shared-gene bridging space and matching Leiden clusters using centroid similarity and marker-gene overlap. Cross-species edges are then selected within the matched cluster pairs, with the number of admitted edges for species pair $(a,b)$ controlled by a divergence-time-aware budget:
\begin{equation}
\pi_{ab} = \exp\!\left(-\frac{t_{ab}}{\tau}\right),
\qquad
B_{ab} =
\left\lfloor
B \cdot
\frac{\pi_{ab}}{\sum_{(a',b')}\pi_{a'b'}}
\right\rfloor,
\end{equation}
where $t_{ab}$ is the estimated divergence time, $\tau$ is a temperature parameter, and $B$ is the total cross-species edge budget. For the Human benchmark, no phylogeny branch is instantiated, i.e., $\mathcal{E}_{\text{Phy}}=\varnothing$. The full edge-scoring and filtering procedure is provided in \appref{app:phylo_details}.

\paragraph{Graph assembly.}
Finally, the three branches are merged by set union and explicit symmetrization:
\begin{equation}
\mathcal{E}=\operatorname{sym}\!\left(
\mathcal{E}_{\text{Align}} \cup
\mathcal{E}_{\text{Onto}} \cup
\mathcal{E}_{\text{Phy}}
\right),
\qquad
A_{ij} = \mathbb{I}\left[(i,j) \in \mathcal{E}\right].
\end{equation}
We use union rather than intersection because the three branches encode distinct relational axes: expression proximity, ontology-level semantic relatedness, and evolutionary conservation. The final graph is kept unweighted, allowing attention-based backbones such as GAT to learn neighbor importance during training.

\subsection{Dual-View Semantic Fusion}
\label{subsec:augmentation}

PCA features capture high-resolution expression variation but lack explicit functional semantics. DOGMA therefore constructs a species-specific GO feature vector $\mathbf{z}_i$ by aggregating expression over genes annotated to selected GO terms, then concatenates the statistical and semantic views. In the main setting, we use $D_{go}=200$. The GO-dimensionality selection experiment and coordinate audit are reported in \appref{app:go200_sensitivity}.
\begin{equation}
\mathbf{H}_i = [\mathbf{x}^{pca}_i \parallel \mathbf{z}_i].
\end{equation}
The resulting representation provides both local expression information and conserved functional anchors for downstream message passing.

\section{Experiments}
To evaluate whether DOGMA improves single-cell analysis through data-centric structural construction, we organize the experiments around five questions: \textbf{Q1: Effectiveness}: Does DOGMA learn a broadly transferable representation across supervised annotation, clustering, and strict zero-shot cell-type evaluation? \textbf{Q2: Fair Comparison}: Does DOGMA retain its strict zero-shot advantage against stronger prior-augmented baselines? \textbf{Q3: Attribution}: Which biological priors and data-construction modules drive DOGMA's performance? \textbf{Q4: Robustness}: Is DOGMA robust when biological priors are missing or noisy, when the ontology threshold is perturbed, and when training data are limited? \textbf{Q5: Efficiency}: Does DOGMA offer practical computational efficiency in time and memory?

\subsection{Cross-Task Generalization}
\label{subsec:cross_task_generalization}

To answer \textbf{Q1: Effectiveness}, we evaluate whether DOGMA transfers across supervised annotation, zero-shot evaluation, and clustering. We group the baselines into three families: representation-centric encoders such as scGPT and scCello, alignment and neighborhood methods such as KNN, MNN, and SATURN, and graph-structured methods such as scPriorGraph and scMoGNN. The first learns embeddings without DOGMA-style prior-guided topology, the second transfers information through local geometry or cross-domain alignment, and the third uses graph structure but still relies on heuristic or molecular-prior neighborhoods rather than multi-level biological priors.

\tabref{tab:main_benchmark} shows that DOGMA is strongest or near-strongest across settings, with especially clear gains on metadata prediction. Against scGPT, DOGMA improves Multi cell-type accuracy from 0.9040 to 0.9333, while scGPT remains slightly better on Brain cell-type annotation (0.9817 vs. 0.9744), likely reflecting its large-scale pretraining. Under strict zero-shot cell-type evaluation, DOGMA improves ARI across all three benchmarks and remains highly competitive for clustering. We use a label-strict transductive zero-shot protocol: unseen cell-type labels are excluded from training, classifier outputs, prototypes, and ontology edges.

\begin{table*}[!t]
\centering
\caption{\textbf{Main Performance Benchmark.} Classification accuracy for Cell Type, Development Stage, and metadata-average prediction across three biological datasets. Metadata Avg averages Cell Type, Development Stage, Sex, and Tissue. Main-table cells report means. Complete 95\% CI half-widths are provided in \apptabref{tab:main_benchmark_full_ci}. We highlight the \hlfirst{\best{best}} and \hlsecond{\second{second best}} results.}
\vspace{-0.4em}
\label{tab:main_benchmark}
\renewcommand\tabcolsep{5.5pt}
\renewcommand\arraystretch{1.15}
\resizebox{\linewidth}{!}{
\begin{tabular}{B|M|ccc|ccc|ccc}
\Xhline{1.2pt}
\rowcolor{tableheadergray}
\tworowhead{Baseline} & \tworowhead{Method} & \multicolumn{3}{c|}{\textbf{Cell Type}} & \multicolumn{3}{c|}{\textbf{Dev. Stage}} & \multicolumn{3}{c}{\textbf{Metadata Avg}} \\
\rowcolor{tableheadergray}
& & \textbf{Brain} & \textbf{Human} & \textbf{Multi} & \textbf{Brain} & \textbf{Human} & \textbf{Multi} & \textbf{Brain} & \textbf{Human} & \textbf{Multi} \\
\Xhline{1.2pt}

\alignmentrowshift
\multirow{3}{*}{\baselinecell{Alignment}{statistical}}
& KNN & 0.9551 & 0.8881 & 0.9029 & 0.8913 & 0.9094 & 0.8505 & 0.9219 & \hlsecond{\second{0.9105}} & 0.9124 \\
& MNN & 0.9611 & 0.8795 & 0.9204 & \hlsecond{\second{0.9077}} & \hlsecond{\second{0.9114}} & \hlsecond{\second{0.8641}} & \hlsecond{\second{0.9304}} & 0.9064 & \hlsecond{\second{0.9245}} \\
& SATURN & 0.9646 & \hlsecond{\second{0.8905}} & \hlfirst{\best{0.9537}} & 0.8174 & 0.7327 & 0.7846 & 0.8663 & 0.8122 & 0.8977 \\

\midrule

\multirow{2}{*}{\baselinecell{Graph}{structure}}
& scPriorGraph & 0.8957 & 0.7500 & 0.6574 & 0.8171 & 0.5701 & 0.7530 & 0.8214 & 0.6809 & 0.8139 \\
& scMoGNN & 0.7691 & 0.5532 & 0.6783 & 0.7988 & 0.8025 & 0.7121 & 0.7783 & 0.7747 & 0.8016 \\

\midrule

\multirow{2}{*}{\baselinecell{Representation}{embedding}}
& scCello & 0.9555 & 0.8200 & 0.8869 & 0.8176 & 0.6054 & 0.7882 & 0.8507 & 0.7179 & 0.8734 \\
& scGPT & \hlfirst{\best{0.9817}} & 0.8643 & 0.9040 & 0.8753 & 0.8648 & 0.8582 & 0.9191 & 0.8782 & 0.9223 \\

\midrule

\dogmarowshift
\baselinecell{DOGMA}{ours}
& \textbf{DOGMA} & \hlsecond{\second{0.9744}} & \hlfirst{\best{0.8983}} & \hlsecond{\second{0.9333}} & \hlfirst{\best{0.9355}} & \hlfirst{\best{0.9414}} & \hlfirst{\best{0.8759}} & \hlfirst{\best{0.9373}} & \hlfirst{\best{0.9386}} & \hlfirst{\best{0.9319}} \\
\Xhline{1.2pt}
\end{tabular}
}
\vspace{-0.3em}
\end{table*}

\begin{table*}[!t]
\centering
\caption{\textbf{Strict Zero-Shot Cell-Type Evaluation and Clustering Benchmark.} Strict zero-shot columns report ARI from seen-class prototype assignments evaluated only on unseen cell-type cells; clustering columns report ARI and AMI in separate subcolumns. Main-table cells report means. CI statistics are reported in Appendix Tables~\ref{tab:zeroshot_full_ci} and~\ref{tab:clustering_full_ci}. We highlight the \hlfirst{\best{best}} and \hlsecond{\second{second best}} results within each column.}
\vspace{-0.4em}
\label{tab:zeroshot_clustering}
\renewcommand\tabcolsep{6pt}
\renewcommand\arraystretch{1.12}
\resizebox{\linewidth}{!}{
\begin{tabular}{B|M|NNN|NNN|NNN}
\Xhline{1.2pt}
\rowcolor{tableheadergray}
\tworowhead{Baseline} & \tworowhead{Method} & \multicolumn{3}{c|}{\textbf{Strict Zero-Shot}} & \multicolumn{3}{c|}{\textbf{Clustering ARI}} & \multicolumn{3}{c}{\textbf{Clustering AMI}} \\
\rowcolor{tableheadergray}
& & \textbf{Brain} & \textbf{Human} & \textbf{Multi} & \textbf{Brain} & \textbf{Human} & \textbf{Multi} & \textbf{Brain} & \textbf{Human} & \textbf{Multi} \\
\Xhline{1.2pt}

\alignmentrowshift
\multirow{3}{*}{\baselinecell{Alignment}{statistical}}
& \onerowmethod{KNN} & \hlsecond{\second{0.5723}} & 0.6265 & 0.5276 & 0.2476 & \hlsecond{\second{0.4747}} & 0.3652 & 0.6111 & 0.7193 & 0.6309 \\
& \onerowmethod{MNN} & 0.5663 & \hlsecond{\second{0.6337}} & 0.4817 & 0.3462 & 0.3682 & 0.3489 & 0.6990 & 0.7266 & 0.7295 \\
& \onerowmethod{SATURN} & 0.2511 & 0.5635 & 0.4870 & 0.4400 & 0.4233 & 0.5290 & 0.7447 & \hlsecond{\second{0.7435}} & 0.7642 \\

\midrule

\multirow{2}{*}{\baselinecell{Graph}{structure}}
& scPriorGraph & 0.4029 & 0.4808 & 0.3265 & 0.4997 & 0.4690 & 0.3591 & 0.6361 & 0.6956 & 0.5659 \\
& scMoGNN & 0.5566 & 0.5605 & 0.3498 & 0.4822 & 0.4162 & 0.4307 & 0.7499 & 0.7276 & 0.6950 \\

\midrule

\multirow{2}{*}{\baselinecell{Representation}{embedding}}
& scCello & 0.4095 & 0.4486 & \hlsecond{\second{0.5542}} & 0.4850 & 0.4659 & \hlsecond{\second{0.5408}} & 0.7482 & 0.7223 & \hlfirst{\best{0.7880}} \\
& scGPT & 0.5205 & 0.5062 & 0.4052 & \hlfirst{\best{0.6229}} & 0.4487 & 0.4901 & \hlsecond{\second{0.7806}} & 0.7062 & 0.7362 \\

\midrule

\dogmarowshift
\baselinecell{DOGMA}{ours}
& \textbf{DOGMA} & \hlfirst{\best{0.5865}} & \hlfirst{\best{0.6638}} & \hlfirst{\best{0.5824}} & \hlsecond{\second{0.5323}} & \hlfirst{\best{0.4767}} & \hlfirst{\best{0.5624}} & \hlfirst{\best{0.7828}} & \hlfirst{\best{0.7438}} & \hlsecond{\second{0.7691}} \\
\Xhline{1.2pt}
\end{tabular}
}
\vspace{-0.8em}
\end{table*}

\subsection{Prior-Augmented Strict Zero-Shot Evaluation}
\label{subsec:prior_augmented_transfer}

\begin{figure*}[t]
    \centering
    \includegraphics[width=0.94\textwidth]{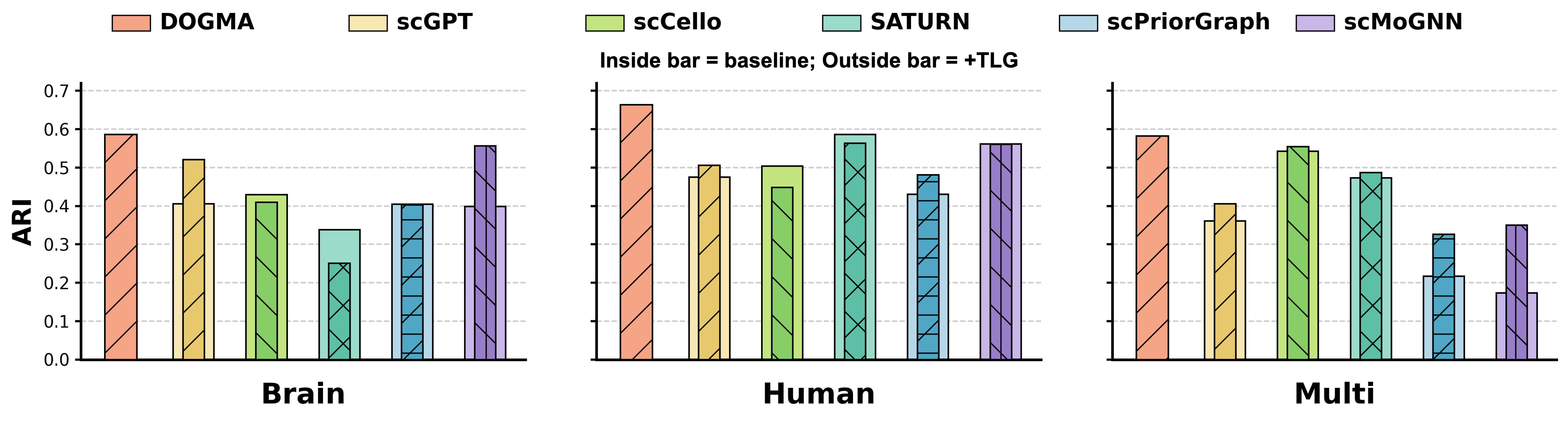}
    \vspace{-0.7em}
    \caption{\textbf{Prior-augmented strict zero-shot evaluation (Q2: Fair Comparison).} Strict zero-shot cell-type ARI on Brain, Human, and Multi. DOGMA is shown as the main-reference method, while competing baselines are evaluated with a seen-only TrainLabelGraph. Baseline-only ARI is shown as the paired reference bar.}
    \label{fig:tlg_transfer}
\end{figure*}

To answer \textbf{Q2: Fair Comparison}, we further test whether DOGMA's strict zero-shot advantage remains when competing methods are strengthened with an additional seen-label prior. For each baseline, we construct a TrainLabelGraph (TLG) using only ontology-derived edges among seen cell types, ensuring that unseen labels are never used for graph construction. This setting gives baselines explicit access to a stronger prior while preserving label-leakage safety. As shown in \figref{fig:tlg_transfer}, DOGMA remains the strongest method by ARI across all three strict zero-shot benchmarks; the paired numerical audit is provided in \apptabref{tab:tlg_strict_zeroshot}. Adding TLG improves some baselines, such as SATURN on Brain and Human, but does not close the gap to DOGMA under the primary strict zero-shot ARI metric. For scGPT and scCello, we keep their native encoders unchanged and attach TLG only after frozen embedding extraction by merging a label-free base graph with the seen-only TLG in a shared graph adapter. Unseen labels are reserved exclusively for final evaluation.

\subsection{Prior Contribution}
\label{subsec:prior_contribution}

To answer \textbf{Q3: Attribution}, we isolate DOGMA's data-construction modules through targeted ablations. \tabref{tab:ablation} shows that the performance gain is not driven by mere heuristic edge source, but by the coordinated use of topology-level and feature-level biological priors. Relative to the main benchmark DOGMA reference, removing all biological priors lowers zero-shot ARI by 0.0960, 0.2481, and 0.1885 on Brain, Human, and Multi, respectively, and also weakens clustering ARI by 0.2423, 0.0894, and 0.1283. These reference-relative gaps indicate that prior-guided graph construction is especially important under cell-type shift, because expression similarity alone often cannot distinguish biologically meaningful neighbors from spurious ones.

The ablations also separate the roles of different biological priors. Cell Ontology provides the strongest topology-level constraint, and removing it sharply degrades strict zero-shot ARI and clustering, especially on Brain and Multi. Phylogeny mainly contributes to cross-species settings. Gene Ontology strengthens annotation and zero-shot evaluation by adding functional feature semantics. Leiden clustering remains unchanged because this evaluator only consumes graph topology.

\begin{table*}[t]
    \centering
    \caption{\textbf{Prior Contribution Ablation (Q3: Attribution).} Impact of removing Gene Ontology features, Cell Ontology edges, phylogeny constraints, and all biological priors. The Full row uses the main benchmark DOGMA reference for annotation, strict zero-shot evaluation, and clustering; drops discussed in the text are relative to this main benchmark reference. \textbf{Bold} denotes the best result in each column.}
    \label{tab:ablation}
    \scriptsize
    \setlength{\tabcolsep}{2pt}
    \begin{tabularx}{\textwidth}{l YYY YYY YYY}
        \toprule
        \multirow{3}{*}{\textbf{Method}} &
        \multicolumn{3}{c}{\textbf{Cell Type Annotation (Acc)}} &
        \multicolumn{3}{c}{\textbf{Strict Zero-Shot Cell Type (ARI)}} &
        \multicolumn{3}{c}{\textbf{Clustering (ARI)}} \\
        \cmidrule(lr){2-4} \cmidrule(lr){5-7} \cmidrule(lr){8-10}

         & \textbf{Brain} & \textbf{Human} & \textbf{Multi}
         & \textbf{Brain} & \textbf{Human} & \textbf{Multi}
         & \textbf{Brain} & \textbf{Human} & \textbf{Multi} \\
        \midrule

        w/o Gene Ontology & 0.8076 & 0.8727 & 0.9044 & 0.4836 & 0.4426 & 0.4049 & 0.5323 & 0.4767 & 0.5624 \\
        w/o Cell Ontology & 0.8113 & 0.8575 & 0.9207 & 0.4954 & 0.4192 & 0.3938 & 0.2597 & 0.4097 & 0.5278 \\
        w/o Phylogeny & 0.8179 & 0.8983 & 0.9201 & 0.5042 & 0.6638 & 0.3868 & 0.3935 & 0.4767 & 0.5594 \\
        w/o All Priors & 0.8165 & 0.8623 & 0.9155 & 0.4905 & 0.4157 & 0.3939 & 0.2900 & 0.3873 & 0.4341 \\

        \midrule
        \rowcolor{lightcyan}
        \textbf{DOGMA (Full)} & \textbf{0.9744} & \textbf{0.8983} & \textbf{0.9333} & \textbf{0.5865} & \textbf{0.6638} & \textbf{0.5824} & \textbf{0.5323} & \textbf{0.4767} & \textbf{0.5624} \\
        \bottomrule
    \end{tabularx}
\end{table*}

\subsection{Robustness to Imperfect Priors and Limited Data}
\label{subsec:robustness}

To answer \textbf{Q4: Robustness}, we test DOGMA under imperfect data-side conditions. We first remove increasing fractions of ontology-derived edges to evaluate how much strict zero-shot performance depends on the ontology prior. \figpartref{fig:robustness_efficiency}{a} shows that even with 80\% ontology-edge dropout, DOGMA retains 94.8\%, 95.6\%, and 88.9\% of full-prior zero-shot ARI on Brain, Human, and Multi.

We then vary the available training data ratio to evaluate whether DOGMA degrades gracefully in data-scarce regimes. As shown in \figpartref{fig:robustness_efficiency}{b}, DOGMA maintains strong metadata-average accuracy as the available training data decreases from 100\% to 10\% on the Multi benchmark. Across this practical low-to-full supervision range, DOGMA remains the strongest evaluated method and shows a smaller accuracy drop than other baselines; the numerical values are reported in \apptabref{tab:data_efficiency_multi}. The ontology-threshold sensitivity audit is reported in \appref{app:hyperparameter_sensitivity}.

\subsection{Computational Practicality}
\label{subsec:computational_practicality}

\begin{figure*}[t]
    \centering
    \begin{subfigure}[t]{0.315\textwidth}
        \centering
        \includegraphics[width=\linewidth]{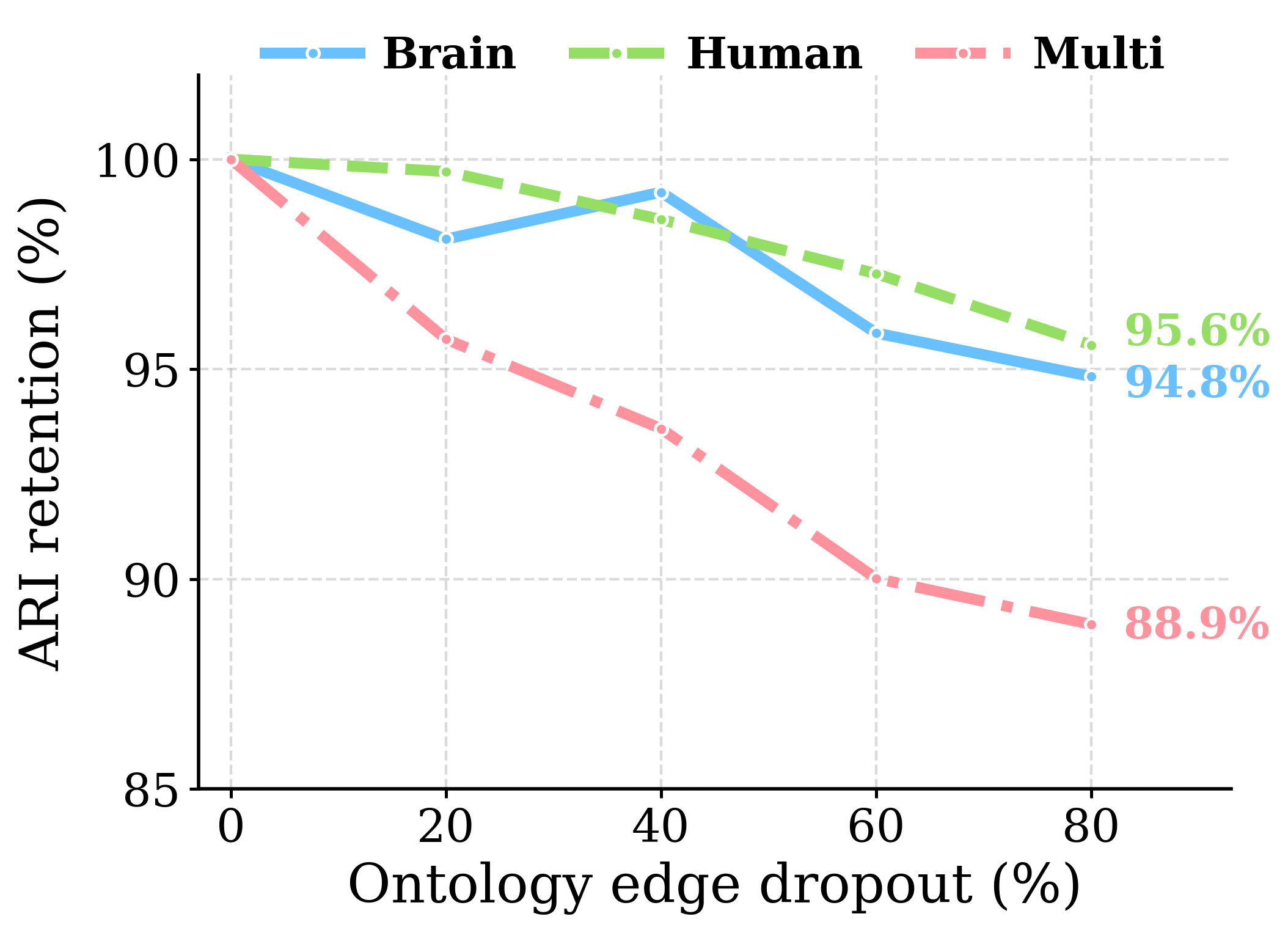}
        \caption{Ontology missingness}
        \label{fig:ontology_missingness}
    \end{subfigure}
    \hfill
    \begin{subfigure}[t]{0.315\textwidth}
        \centering
        \includegraphics[width=\linewidth,trim=8bp 8bp 8bp 8bp,clip]{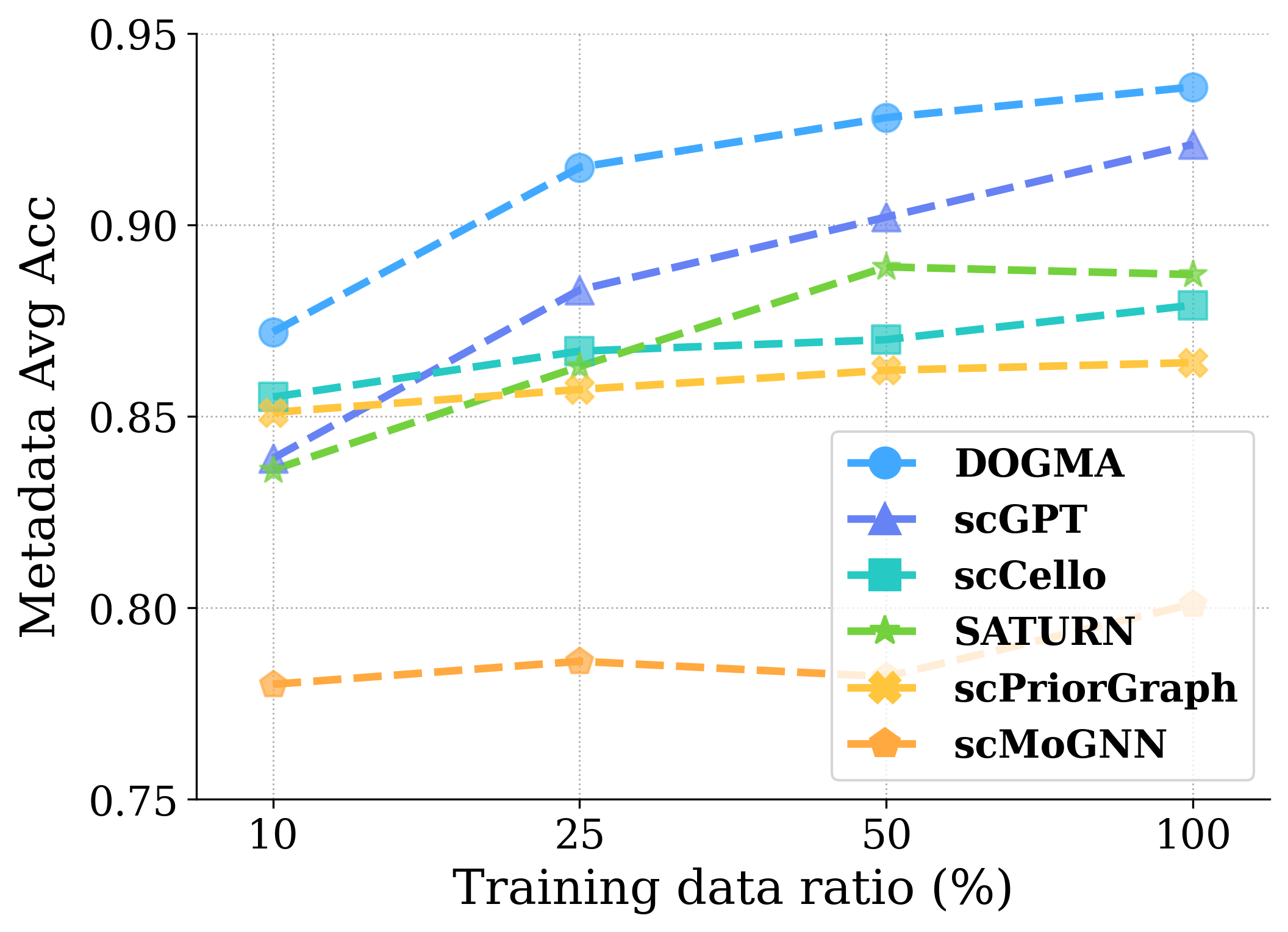}
        \caption{Data efficiency}
        \label{fig:data_efficiency}
    \end{subfigure}
    \hfill
    \begin{subfigure}[t]{0.315\textwidth}
        \centering
        \includegraphics[width=\linewidth,trim=8bp 8bp 8bp 8bp,clip]{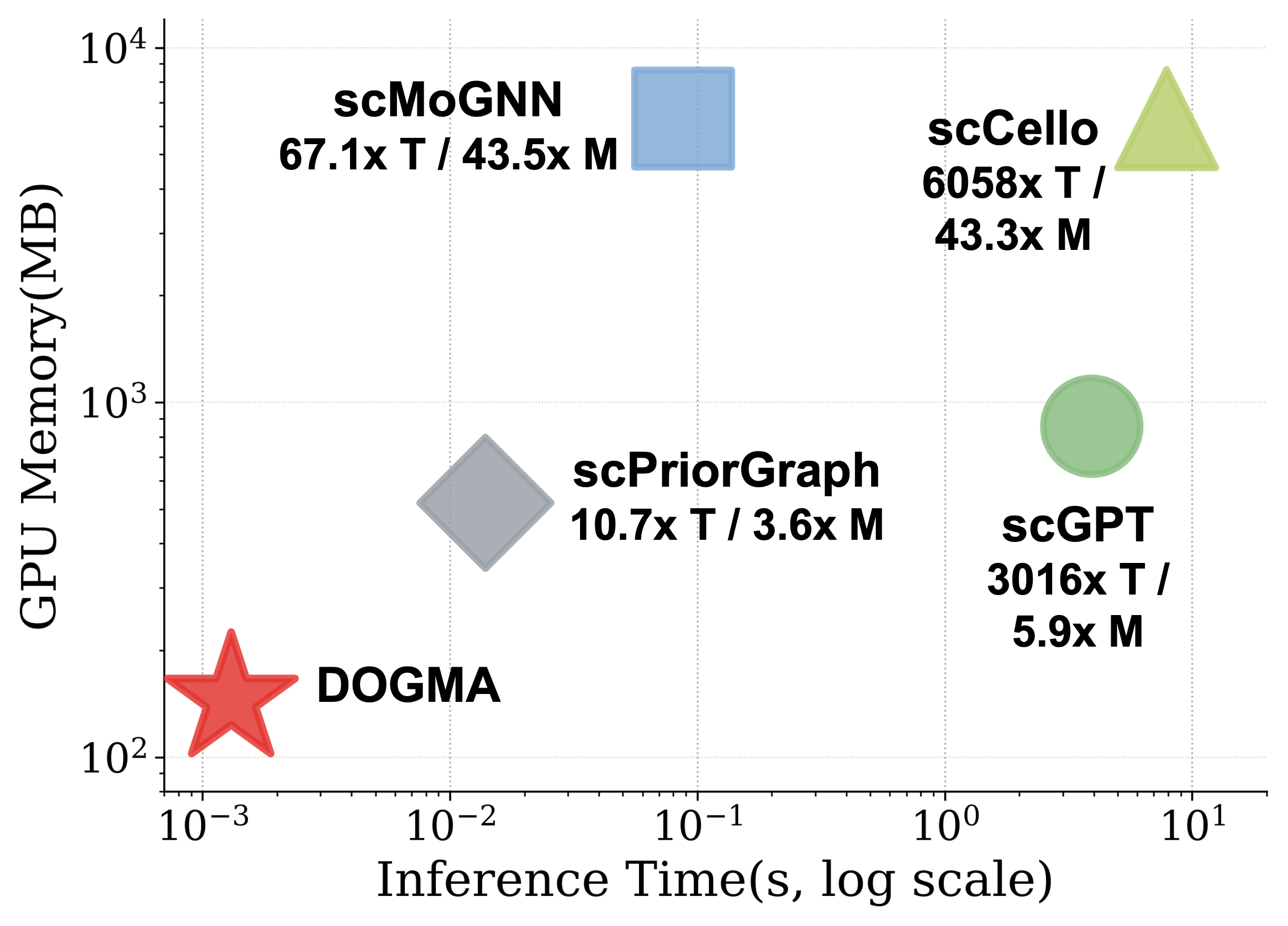}
        \caption{Downstream efficiency}
        \label{fig:q5_efficiency}
    \end{subfigure}
    \vspace{-0.7em}
    \caption{\textbf{Robustness, data efficiency, and downstream efficiency.} (a) Strict zero-shot ARI retention under increasing ontology-edge dropout. (b) Multi benchmark metadata-average accuracy as the training data ratio decreases from 100\% to 10\%. (c) Downstream inference time and reserved GPU memory. DOGMA reports average inference time and the observed GPU-memory range.}
    \label{fig:robustness_efficiency}
\end{figure*}

To answer \textbf{Q5: Efficiency}, we compare the prepared-forward downstream evaluation time and GPU memory footprint of DOGMA against representative high-performing baselines on the graph3 100\% setting. This scope isolates downstream evaluator cost and excludes graph construction, hyperparameter search, model downloading, and warm-up overheads, as detailed in \appref{app:downstream_resource_usage}. As shown in \figpartref{fig:robustness_efficiency}{c}, DOGMA occupies the low-time, low-memory region while preserving strong metadata prediction and strict zero-shot performance. Its downstream inference takes only 0.0013 seconds in this window, while the compared baselines are roughly 11 to 6,000 times slower.

The memory profile shows the same practical advantage within this downstream window. DOGMA reserves 110--182 MB of GPU memory, with a 146 MB average used for ratio calculations, whereas scMoGNN and scCello reserve more than 6.3 GB. This corresponds to about a 43-fold reduction relative to the heaviest graph-structured and representation-centric baselines. These results suggest that DOGMA's data-centric design shifts repeated downstream computation away from large-model inference and toward reusable prior-guided data construction.

\section{Conclusion}
In this paper, we address a data-structure gap in single-cell transcriptomics. Raw scRNA-seq data are noisy and sparse, yet existing input representations either omit explicit cell-cell topology or derive it from heuristic signals with limited biological constraints. To close this gap, we propose \textbf{DOGMA}, a data-centric, knowledge-anchored framework that integrates Gene Ontology, Cell Ontology, and phylogenetic priors to construct biologically consistent cellular networks.

Evaluations show that DOGMA achieves strong performance across supervised annotation, zero-shot cell-type identification, and clustering, with impressive gains in metadata prediction and zero-shot settings. It also remains effective under prior perturbations and data-scarce scenarios while requiring substantially lower downstream inference cost than larger representation-centric and graph-structured baselines, demonstrating that prior-guided graph is both accurate and practical.

\bibliographystyle{plainnat}
\bibliography{example_paper}

\newpage
\appendix
\onecolumn

\section{Limitations}
\label{app:limitations}

DOGMA depends on the quality and coverage of external biological knowledge. Errors, missing terms, inconsistent granularity, or outdated relationships in Cell Ontology, Gene Ontology, source annotations, and reference atlases can propagate into node features, ontology edges, and downstream labels. This dependence is especially relevant for rare cell populations, disease states, or datasets whose annotations do not align cleanly with the ontology terms used during graph construction.

The cross-species priors used by DOGMA are necessarily incomplete abstractions of biology. Phylogenetic distance, organ labels, orthologous genes, and shared ontology structure provide useful guidance, but they may not capture lineage-specific regulatory programs, convergent cell states, non-orthologous functional analogs, or context-dependent conservation across tissues and developmental stages. As a result, the learned graph may underconnect biologically corresponding cells or overconnect superficially similar cells when the available priors are coarse or incomplete.

Our empirical evaluation is limited to the selected public benchmarks, tasks, and baselines considered in this paper. Although the results cover multi-species and multi-organ settings, they do not exhaustively test all organisms, tissues, sequencing technologies, atlas construction protocols, disease contexts, or annotation schemes. Broader validation on larger and more diverse single-cell atlases is needed to characterize when prior-guided graph construction is most reliable and when additional domain-specific calibration is required.

\section{GO-Dimension Sensitivity and Cross-Species Coordinate Audit}
\label{app:go200_sensitivity}

This appendix examines whether selecting 200 coverage- and specificity-prioritized GO terms is empirically stable and whether per-species term selection weakens cross-species semantic alignment. The goal is not to establish $K=200$ as a unique optimum, but to test whether GO200 is a stable, compact empirical setting. We evaluate graph3 with a fixed five-seed GraphBest audit protocol; these values are used for within-audit sensitivity comparisons, while the main benchmark tables remain the primary reference scores. For the local-coordinate sweep, \texttt{local\_go0} removes the GO block and uses PCA features only, while \texttt{local\_go50}, \texttt{local\_go100}, and \texttt{local\_go200} retain the current per-species GO-coordinate pipeline truncated to the first 50, 100, or 200 GO columns.

We also include two coordinate controls at $K=200$. The \texttt{global\_shared\_go200} condition recomputes a shared set of 200 GO terms jointly across species, whereas \texttt{permuted\_local\_go200} preserves each species' GO-value distribution but destroys the local GO-column order. These controls directly test whether independent per-species GO coordinates measurably impair cross-species alignment in this benchmark.

\tabref{tab:go200_sensitivity_cls} reports GraphBest cell-type classification accuracy. Classification performance is already close to saturation by $K=100$--$200$: \texttt{local\_go200} has the highest mean, but its margin over \texttt{local\_go100} is small. This supports GO200 as a reasonable saturated setting rather than a statistically unique optimum.

\begin{table}[H]
    \centering
    \caption{\textbf{GO-dimension sensitivity for cell-type classification on graph3.} Results use GraphBest classification accuracy over five seeds. Delta is computed relative to \texttt{local\_go200}.}
    \label{tab:go200_sensitivity_cls}
    \scriptsize
    \setlength{\tabcolsep}{4pt}
    \renewcommand\arraystretch{1.12}
    \resizebox{\linewidth}{!}{
    \begin{tabular}{lccccccc}
    \toprule
    \textbf{Condition} & \textbf{Coord.} & \textbf{GO dim} & \textbf{x dim} & \textbf{Acc.\ mean} & \textbf{Std} & \textbf{95\% CI} & \textbf{$\Delta$} \\
    \midrule
    \texttt{local\_go0} & local & 0 & 50 & 0.9229 & 0.0056 & [0.9180, 0.9278] & -0.0042 \\
    \texttt{local\_go50} & local & 50 & 100 & 0.9245 & 0.0043 & [0.9207, 0.9282] & -0.0026 \\
    \texttt{local\_go100} & local & 100 & 150 & 0.9269 & 0.0053 & [0.9223, 0.9315] & -0.0002 \\
    \texttt{local\_go200} & local & 200 & 250 & 0.9271 & 0.0041 & [0.9235, 0.9307] & 0.0000 \\
    \texttt{global\_shared\_go200} & global & 200 & 250 & 0.9256 & 0.0033 & [0.9227, 0.9285] & -0.0015 \\
    \texttt{permuted\_local\_go200} & permuted & 200 & 250 & 0.9257 & 0.0023 & [0.9237, 0.9278] & -0.0013 \\
    \bottomrule
    \end{tabular}
    }
    \vspace{-0.5em}
\end{table}

\tabref{tab:go200_sensitivity_zs} reports GraphBest zero-shot ARI for the same conditions. The \texttt{local\_go200} audit mean is 0.5910, close to the main benchmark Multi strict zero-shot ARI of 0.5824 and within its reported 95\% CI half-width of 0.0316. Zero-shot transfer is more sensitive to the GO dimensionality: \texttt{local\_go200} improves over \texttt{local\_go0}, \texttt{local\_go50}, and \texttt{local\_go100} by 0.0598, 0.0470, and 0.0582, respectively. The shared-coordinate and permuted controls do not improve over the original local GO200 representation, so we do not observe a measurable loss from per-species GO-coordinate selection in this benchmark.

\begin{table}[H]
    \centering
    \caption{\textbf{GO-dimension sensitivity for zero-shot transfer on graph3.} Results use GraphBest zero-shot ARI over five seeds. Delta is computed relative to \texttt{local\_go200}.}
    \label{tab:go200_sensitivity_zs}
    \scriptsize
    \setlength{\tabcolsep}{4pt}
    \renewcommand\arraystretch{1.12}
    \resizebox{\linewidth}{!}{
    \begin{tabular}{lccccccc}
    \toprule
    \textbf{Condition} & \textbf{Coord.} & \textbf{GO dim} & \textbf{x dim} & \textbf{ARI mean} & \textbf{Std} & \textbf{95\% CI} & \textbf{$\Delta$} \\
    \midrule
    \texttt{local\_go0} & local & 0 & 50 & 0.5313 & 0.0653 & [0.4740, 0.5885] & -0.0598 \\
    \texttt{local\_go50} & local & 50 & 100 & 0.5441 & 0.0411 & [0.5080, 0.5801] & -0.0470 \\
    \texttt{local\_go100} & local & 100 & 150 & 0.5329 & 0.0530 & [0.4864, 0.5793] & -0.0582 \\
    \texttt{local\_go200} & local & 200 & 250 & 0.5910 & 0.0453 & [0.5514, 0.6307] & 0.0000 \\
    \texttt{global\_shared\_go200} & global & 200 & 250 & 0.5560 & 0.0449 & [0.5167, 0.5954] & -0.0350 \\
    \texttt{permuted\_local\_go200} & permuted & 200 & 250 & 0.5608 & 0.0434 & [0.5228, 0.5989] & -0.0302 \\
    \bottomrule
    \end{tabular}
    }
    \vspace{-0.5em}
\end{table}

Together, these results support GO200 as an empirical setting selected after a $K \in \{0,50,100,200\}$ sensitivity analysis. Classification accuracy is nearly saturated by $K=100$--$200$, whereas zero-shot ARI benefits more clearly from the fuller GO block.

\section{Additional Benchmark Tables}
\label{app:additional_benchmark_tables}

\begin{table}[H]
\centering
\caption{\textbf{Complete Main Performance Benchmark with 95\% CI.} Cells report mean with 95\% CI half-width for Cell Type, Development Stage, and metadata-average prediction across three biological datasets. Metadata Avg averages Cell Type, Development Stage, Sex, and Tissue. We highlight the \hlfirst{\best{best}} and \hlsecond{\second{second best}} results within each column.}
\vspace{0.2em}
\label{tab:main_benchmark_full_ci}
\renewcommand\tabcolsep{3.5pt}
\renewcommand\arraystretch{1.15}
\scriptsize
\resizebox{\linewidth}{!}{
\begin{tabular}{B|M|ccc|ccc|ccc}
\Xhline{1.2pt}
\rowcolor{tableheadergray}
\tworowhead{Baseline} & \tworowhead{Method} & \multicolumn{3}{c|}{\textbf{Cell Type}} & \multicolumn{3}{c|}{\textbf{Dev. Stage}} & \multicolumn{3}{c}{\textbf{Metadata Avg}} \\
\rowcolor{tableheadergray}
& & \textbf{Brain} & \textbf{Human} & \textbf{Multi} & \textbf{Brain} & \textbf{Human} & \textbf{Multi} & \textbf{Brain} & \textbf{Human} & \textbf{Multi} \\
\Xhline{1.2pt}

\multirow{3}{*}{\baselinecell{Alignment}{statistical}}
& KNN & \norm{0.9551}{0.0032} & \norm{0.8881}{0.0015} & \norm{0.9029}{0.0038} & \norm{0.8913}{0.0058} & \norm{0.9094}{0.0023} & \norm{0.8505}{0.0033} & \norm{0.9219}{0.0020} & \hlsecond{\second{\norm{0.9105}{0.0011}}} & \norm{0.9124}{0.0016} \\
& MNN & \norm{0.9611}{0.0020} & \norm{0.8795}{0.0015} & \norm{0.9204}{0.0013} & \hlsecond{\second{\norm{0.9077}{0.0023}}} & \hlsecond{\second{\norm{0.9114}{0.0034}}} & \hlsecond{\second{\norm{0.8641}{0.0025}}} & \hlsecond{\second{\norm{0.9304}{0.0012}}} & \norm{0.9064}{0.0012} & \hlsecond{\second{\norm{0.9245}{0.0009}}} \\
& SATURN & \norm{0.9646}{0.0018} & \hlsecond{\second{\norm{0.8905}{0.0033}}} & \hlfirst{\best{\norm{0.9537}{0.0022}}} & \norm{0.8174}{0.0007} & \norm{0.7327}{0.0032} & \norm{0.7846}{0.0026} & \norm{0.8663}{0.0020} & \norm{0.8122}{0.0017} & \norm{0.8977}{0.0021} \\

\midrule

\multirow{2}{*}{\baselinecell{Graph}{structure}}
& scPriorGraph & \norm{0.8957}{0.0014} & \norm{0.7500}{0.0029} & \norm{0.6574}{0.0076} & \norm{0.8171}{0.0022} & \norm{0.5701}{0.0053} & \norm{0.7530}{0.0046} & \norm{0.8214}{0.0010} & \norm{0.6809}{0.0019} & \norm{0.8139}{0.0023} \\
& scMoGNN & \norm{0.7691}{0.0323} & \norm{0.5532}{0.0146} & \norm{0.6783}{0.0120} & \norm{0.7988}{0.0075} & \norm{0.8025}{0.0294} & \norm{0.7121}{0.0013} & \norm{0.7783}{0.0094} & \norm{0.7747}{0.0100} & \norm{0.8016}{0.0036} \\

\midrule

\multirow{2}{*}{\baselinecell{Representation}{embedding}}
& scCello & \norm{0.9555}{0.0013} & \norm{0.8200}{0.0010} & \norm{0.8869}{0.0011} & \norm{0.8176}{0.0015} & \norm{0.6054}{0.0031} & \norm{0.7882}{0.0014} & \norm{0.8507}{0.0008} & \norm{0.7179}{0.0012} & \norm{0.8734}{0.0008} \\
& scGPT & \hlfirst{\best{\norm{0.9817}{0.0028}}} & \norm{0.8643}{0.0033} & \norm{0.9040}{0.0036} & \norm{0.8753}{0.0040} & \norm{0.8648}{0.0066} & \norm{0.8582}{0.0032} & \norm{0.9191}{0.0016} & \norm{0.8782}{0.0029} & \norm{0.9223}{0.0013} \\

\midrule

\baselinecell{DOGMA}{ours}
& \textbf{DOGMA} & \hlsecond{\second{\norm{0.9744}{0.0010}}} & \hlfirst{\best{\norm{0.8983}{0.0026}}} & \hlsecond{\second{\norm{0.9333}{0.0019}}} & \hlfirst{\best{\norm{0.9355}{0.0008}}} & \hlfirst{\best{\norm{0.9414}{0.0018}}} & \hlfirst{\best{\norm{0.8759}{0.0000}}} & \hlfirst{\best{\norm{0.9373}{0.0008}}} & \hlfirst{\best{\norm{0.9386}{0.0013}}} & \hlfirst{\best{\norm{0.9319}{0.0007}}} \\
\Xhline{1.2pt}
\end{tabular}
}
\vspace{-0.5em}
\end{table}

\begin{table}[H]
\centering
\caption{\textbf{Complete Strict Zero-Shot Benchmark with 95\% CI.} Cells report mean ARI with 95\% CI half-width for Brain, Human, and Multi. We highlight the \hlfirst{\best{best}} and \hlsecond{\second{second best}} results within each column.}
\vspace{0.2em}
\label{tab:zeroshot_full_ci}
\renewcommand\tabcolsep{4.5pt}
\renewcommand\arraystretch{1.15}
\scriptsize
\resizebox{\linewidth}{!}{
\begin{tabular}{B|M|ccc}
\Xhline{1.2pt}
\rowcolor{tableheadergray}
\tworowhead{Baseline} & \tworowhead{Method} & \multicolumn{3}{c}{\textbf{Strict Zero-Shot ARI}} \\
\rowcolor{tableheadergray}
& & \textbf{Brain} & \textbf{Human} & \textbf{Multi} \\
\Xhline{1.2pt}

\multirow{3}{*}{\baselinecell{Alignment}{statistical}}
& \onerowmethod{KNN} & \hlsecond{\second{\norm{0.5723}{0.0118}}} & \norm{0.6265}{0.0264} & \norm{0.5276}{0.0303} \\
& \onerowmethod{MNN} & \norm{0.5663}{0.0154} & \hlsecond{\second{\norm{0.6337}{0.0424}}} & \norm{0.4817}{0.0204} \\
& \onerowmethod{SATURN} & \norm{0.2511}{0.0312} & \norm{0.5635}{0.0061} & \norm{0.4870}{0.0060} \\

\midrule

\multirow{2}{*}{\baselinecell{Graph}{structure}}
& scPriorGraph & \norm{0.4029}{0.0084} & \norm{0.4808}{0.0079} & \norm{0.3265}{0.0113} \\
& scMoGNN & \norm{0.5566}{0.0207} & \norm{0.5605}{0.0100} & \norm{0.3498}{0.0332} \\

\midrule

\multirow{2}{*}{\baselinecell{Representation}{embedding}}
& scCello & \norm{0.4095}{0.0352} & \norm{0.4486}{0.0093} & \hlsecond{\second{\norm{0.5542}{0.0415}}} \\
& scGPT & \norm{0.5205}{0.0227} & \norm{0.5062}{0.0129} & \norm{0.4052}{0.0189} \\

\midrule

\dogmarowshift
\baselinecell{DOGMA}{ours}
& \textbf{DOGMA} & \hlfirst{\best{\norm{0.5865}{0.0450}}} & \hlfirst{\best{\norm{0.6638}{0.0286}}} & \hlfirst{\best{\norm{0.5824}{0.0316}}} \\
\Xhline{1.2pt}
\end{tabular}
}
\vspace{-0.5em}
\end{table}

\begin{table}[H]
\centering
\caption{\textbf{Complete Clustering Benchmark with 95\% CI.} Cells report mean with 95\% CI half-width for ARI and AMI across Brain, Human, and Multi. We highlight the \hlfirst{\best{best}} and \hlsecond{\second{second best}} results within each column.}
\vspace{0.2em}
\label{tab:clustering_full_ci}
\renewcommand\tabcolsep{4pt}
\renewcommand\arraystretch{1.15}
\scriptsize
\resizebox{\linewidth}{!}{
\begin{tabular}{B|M|ccc|ccc}
\Xhline{1.2pt}
\rowcolor{tableheadergray}
\tworowhead{Baseline} & \tworowhead{Method} & \multicolumn{3}{c|}{\textbf{Clustering ARI}} & \multicolumn{3}{c}{\textbf{Clustering AMI}} \\
\rowcolor{tableheadergray}
& & \textbf{Brain} & \textbf{Human} & \textbf{Multi} & \textbf{Brain} & \textbf{Human} & \textbf{Multi} \\
\Xhline{1.2pt}

\multirow{3}{*}{\baselinecell{Alignment}{statistical}}
& \onerowmethod{KNN} & \norm{0.2476}{0.0047} & \hlsecond{\second{\norm{0.4747}{0.0067}}} & \norm{0.3652}{0.0070} & \norm{0.6111}{0.0059} & \norm{0.7193}{0.0034} & \norm{0.6309}{0.0050} \\
& \onerowmethod{MNN} & \norm{0.3462}{0.0040} & \norm{0.3682}{0.0020} & \norm{0.3489}{0.0054} & \norm{0.6990}{0.0041} & \norm{0.7266}{0.0014} & \norm{0.7295}{0.0024} \\
& \onerowmethod{SATURN} & \norm{0.4400}{0.0020} & \norm{0.4233}{0.0047} & \norm{0.5290}{0.0070} & \norm{0.7447}{0.0014} & \hlsecond{\second{\norm{0.7435}{0.0045}}} & \norm{0.7642}{0.0014} \\

\midrule

\multirow{2}{*}{\baselinecell{Graph}{structure}}
& scPriorGraph & \norm{0.4997}{0.0000} & \norm{0.4690}{0.0000} & \norm{0.3591}{0.0000} & \norm{0.6361}{0.0000} & \norm{0.6956}{0.0000} & \norm{0.5659}{0.0000} \\
& scMoGNN & \norm{0.4822}{0.0117} & \norm{0.4162}{0.0079} & \norm{0.4307}{0.0053} & \norm{0.7499}{0.0024} & \norm{0.7276}{0.0038} & \norm{0.6950}{0.0032} \\

\midrule

\multirow{2}{*}{\baselinecell{Representation}{embedding}}
& scCello & \norm{0.4850}{0.0158} & \norm{0.4659}{0.0236} & \hlsecond{\second{\norm{0.5408}{0.0255}}} & \norm{0.7482}{0.0040} & \norm{0.7223}{0.0073} & \hlfirst{\best{\norm{0.7880}{0.0074}}} \\
& scGPT & \hlfirst{\best{\norm{0.6229}{0.0014}}} & \norm{0.4487}{0.0065} & \norm{0.4901}{0.0115} & \hlsecond{\second{\norm{0.7806}{0.0007}}} & \norm{0.7062}{0.0022} & \norm{0.7362}{0.0033} \\

\midrule

\dogmarowshift
\baselinecell{DOGMA}{ours}
& \textbf{DOGMA} & \hlsecond{\second{\norm{0.5323}{0.0007}}} & \hlfirst{\best{\norm{0.4767}{0.0015}}} & \hlfirst{\best{\norm{0.5624}{0.0134}}} & \hlfirst{\best{\norm{0.7828}{0.0002}}} & \hlfirst{\best{\norm{0.7438}{0.0016}}} & \hlsecond{\second{\norm{0.7691}{0.0023}}} \\
\Xhline{1.2pt}
\end{tabular}
}
\vspace{-0.5em}
\end{table}

\begin{table}[H]
\centering
\caption{\textbf{Prior-Augmented Strict Zero-Shot Benchmark.} Baseline and TrainLabelGraph (TLG)-augmented strict zero-shot ARI across Brain, Human, and Multi. Baseline ARI values are copied from the main strict zero-shot table; cells report mean with 95\% CI half-width over $n=10$ random seeds. Delta is computed as +TLG ARI minus the main-table baseline ARI.}
\vspace{0.2em}
\label{tab:tlg_strict_zeroshot}
\renewcommand\tabcolsep{5pt}
\renewcommand\arraystretch{1.12}
\scriptsize
\resizebox{\linewidth}{!}{
\begin{tabular}{llcccr}
\toprule
\textbf{Dataset} & \textbf{Method} & \textbf{Baseline ARI} & \textbf{+TLG ARI} & \textbf{Delta} & \textbf{$n$} \\
\midrule
Brain & DOGMA & \norm{0.5865}{0.0450} & -- & -- & 10 \\
Brain & KNN & \norm{0.5723}{0.0118} & \norm{0.5739}{0.0212} & +0.0016 & 10 \\
Brain & MNN & \norm{0.5663}{0.0154} & \norm{0.5575}{0.0122} & -0.0088 & 10 \\
Brain & SATURN & \norm{0.2511}{0.0312} & \norm{0.3378}{0.0096} & +0.0867 & 10 \\
Brain & scPriorGraph & \norm{0.4029}{0.0084} & \norm{0.4047}{0.0092} & +0.0018 & 10 \\
Brain & scMoGNN & \norm{0.5566}{0.0207} & \norm{0.3986}{0.0057} & -0.1580 & 10 \\
Brain & scCello & \norm{0.4095}{0.0352} & \norm{0.4295}{0.0229} & +0.0200 & 10 \\
Brain & scGPT & \norm{0.5205}{0.0227} & \norm{0.4054}{0.0210} & -0.1151 & 10 \\
\midrule
Human & DOGMA & \norm{0.6638}{0.0286} & -- & -- & 10 \\
Human & KNN & \norm{0.6265}{0.0264} & \norm{0.6390}{0.0189} & +0.0125 & 10 \\
Human & MNN & \norm{0.6337}{0.0424} & \norm{0.6224}{0.0423} & -0.0113 & 10 \\
Human & SATURN & \norm{0.5635}{0.0061} & \norm{0.5865}{0.0039} & +0.0230 & 10 \\
Human & scPriorGraph & \norm{0.4808}{0.0079} & \norm{0.4309}{0.0085} & -0.0499 & 10 \\
Human & scMoGNN & \norm{0.5605}{0.0100} & \norm{0.5619}{0.0187} & +0.0014 & 10 \\
Human & scCello & \norm{0.4486}{0.0093} & \norm{0.5038}{0.0084} & +0.0552 & 10 \\
Human & scGPT & \norm{0.5062}{0.0129} & \norm{0.4750}{0.0165} & -0.0312 & 10 \\
\midrule
Multi & DOGMA & \norm{0.5824}{0.0316} & -- & -- & 10 \\
Multi & KNN & \norm{0.5276}{0.0303} & \norm{0.5014}{0.0348} & -0.0262 & 10 \\
Multi & MNN & \norm{0.4817}{0.0204} & \norm{0.4683}{0.0324} & -0.0134 & 10 \\
Multi & SATURN & \norm{0.4870}{0.0060} & \norm{0.4730}{0.0064} & -0.0140 & 10 \\
Multi & scPriorGraph & \norm{0.3265}{0.0113} & \norm{0.2169}{0.0086} & -0.1096 & 10 \\
Multi & scMoGNN & \norm{0.3498}{0.0332} & \norm{0.1731}{0.0074} & -0.1767 & 10 \\
Multi & scCello & \norm{0.5542}{0.0415} & \norm{0.5426}{0.0350} & -0.0116 & 10 \\
Multi & scGPT & \norm{0.4052}{0.0189} & \norm{0.3634}{0.0153} & -0.0418 & 10 \\
\bottomrule
\end{tabular}
}
\vspace{-0.5em}
\end{table}

\begin{table}[H]
\centering
\caption{\textbf{Multi Data-Efficiency Benchmark.} Metadata-average classification accuracy on the Multi benchmark as the available training data ratio changes. Cells report mean with 95\% CI half-width. Metadata Avg averages Cell Type, Development Stage, Sex, and Tissue.}
\label{tab:data_efficiency_multi}
\scriptsize
\setlength{\tabcolsep}{5pt}
\renewcommand\arraystretch{1.12}
\resizebox{\linewidth}{!}{
\begin{tabular}{lcccc}
\toprule
\textbf{Method} & \textbf{10\%} & \textbf{25\%} & \textbf{50\%} & \textbf{100\%} \\
\midrule
DOGMA & \norm{0.8724}{0.0010} & \norm{0.9154}{0.0021} & \norm{0.9278}{0.0008} & \norm{0.9365}{0.0010} \\
scGPT & \norm{0.8390}{0.0033} & \norm{0.8828}{0.0068} & \norm{0.9016}{0.0030} & \norm{0.9207}{0.0044} \\
scCello & \norm{0.8551}{0.0029} & \norm{0.8671}{0.0011} & \norm{0.8703}{0.0022} & \norm{0.8785}{0.0026} \\
SATURN & \norm{0.8363}{0.0064} & \norm{0.8632}{0.0038} & \norm{0.8895}{0.0055} & \norm{0.8874}{0.0036} \\
scPriorGraph & \norm{0.8509}{0.0063} & \norm{0.8569}{0.0030} & \norm{0.8624}{0.0030} & \norm{0.8641}{0.0004} \\
scMoGNN & \norm{0.7802}{0.0098} & \norm{0.7861}{0.0161} & \norm{0.7817}{0.0170} & \norm{0.8014}{0.0185} \\
\bottomrule
\end{tabular}
}
\vspace{-0.5em}
\end{table}
\clearpage

\section{Hyperparameter Sensitivity}
\label{app:hyperparameter_sensitivity}

We first audit sensitivity to ontology-edge missingness by progressively dropping ontology-derived edges and measuring strict zero-shot ARI. The full-prior row is the audit reference for the retention percentages reported in \figpartref{fig:robustness_efficiency}{a}.

\begin{table}[H]
    \centering
    \caption{\textbf{Ontology-edge missingness audit.} Strict zero-shot ARI under increasing ontology-edge dropout. Values in brackets report the 95\% confidence interval.}
    \label{tab:ontology_edge_missingness}
    \scriptsize
    \setlength{\tabcolsep}{5pt}
    \renewcommand\arraystretch{1.12}
    \resizebox{\linewidth}{!}{
    \begin{tabular}{rrccc}
    \toprule
    \textbf{Remaining edges (\%)} & \textbf{Dropout (\%)} & \textbf{Brain ARI} & \textbf{Human ARI} & \textbf{Multi ARI} \\
    \midrule
    100 & 0 & 0.4829 [0.4684, 0.4974] & 0.6125 [0.5844, 0.6405] & 0.5577 [0.5194, 0.5959] \\
    80 & 20 & 0.4737 [0.4496, 0.4979] & 0.6107 [0.5849, 0.6364] & 0.5338 [0.4933, 0.5743] \\
    60 & 40 & 0.4791 [0.4608, 0.4974] & 0.6037 [0.5759, 0.6316] & 0.5219 [0.4791, 0.5648] \\
    40 & 60 & 0.4629 [0.4448, 0.4809] & 0.5958 [0.5699, 0.6218] & 0.5020 [0.4670, 0.5370] \\
    20 & 80 & 0.4579 [0.4394, 0.4763] & 0.5854 [0.5622, 0.6086] & 0.4959 [0.4531, 0.5387] \\
    \bottomrule
    \end{tabular}
    }
    \vspace{-0.5em}
\end{table}

We audit sensitivity to the ontology-distance threshold $\tau_b$ around the selected operating point used in \tabref{tab:zeroshot_clustering}. Here, \texttt{graph1}, \texttt{graph2}, and \texttt{graph3} correspond to Brain, Human, and Multi, respectively; the Human benchmark uses HCAO rather than the standard CL, so the CL threshold audit focuses on \texttt{graph1} and \texttt{graph3}. Starred rows denote the default operating points used in the main table, and their mean ARI and AMI values are synchronized with the main-table DOGMA clustering scores. Although the main operating range caps $\tau_b$ at $2$, we include $\tau_b=3$ as a perturbation audit beyond the default cap.

\begin{table}[H]
    \centering
    \caption{\textbf{Ontology-threshold sensitivity audit.} Sensitivity of clustering ARI and AMI to the ontology-distance threshold $\tau_b$. Starred values mark the default settings used in the main experiments.}
    \label{tab:tau_sensitivity}
    \scriptsize
    \setlength{\tabcolsep}{4pt}
    \renewcommand\arraystretch{1.12}
    \resizebox{\linewidth}{!}{
    \begin{tabular}{llccccc}
    \toprule
    \textbf{Graph} & \textbf{Param} & \textbf{Value} & \textbf{ARI mean} & \textbf{ARI 95\% CI} & \textbf{AMI mean} & \textbf{AMI 95\% CI} \\
    \midrule
    \texttt{graph1} & $\tau_b$ & $1.0000^\ast$ & 0.5323 & [0.5316, 0.5330] & 0.7828 & [0.7826, 0.7830] \\
    \texttt{graph1} & $\tau_b$ & 2.0000 & 0.5068 & [0.5004, 0.5132] & 0.7616 & [0.7567, 0.7664] \\
    \texttt{graph1} & $\tau_b$ & 3.0000 & 0.5150 & [0.5101, 0.5198] & 0.7714 & [0.7684, 0.7744] \\
    \texttt{graph3} & $\tau_b$ & 1.0000 & 0.5474 & [0.5334, 0.5613] & 0.7696 & [0.7673, 0.7719] \\
    \texttt{graph3} & $\tau_b$ & $2.0000^\ast$ & 0.5624 & [0.5490, 0.5758] & 0.7691 & [0.7668, 0.7714] \\
    \texttt{graph3} & $\tau_b$ & 3.0000 & 0.5575 & [0.5479, 0.5672] & 0.7637 & [0.7623, 0.7652] \\
    \bottomrule
    \end{tabular}
    }
    \vspace{-0.5em}
\end{table}
\clearpage

\section{DOGMA Method Details}
\label{app:method_details}

This section expands the compact method description in Section~\ref{sec:methods}. DOGMA consists of data preprocessing, three independent edge-construction branches, graph assembly, and GO-based feature augmentation.

\subsection{Preprocessing and Feature Initialization}
We standardize raw single-cell transcriptomic data from CELLxGENE before graph construction. Genes expressed in fewer than 3 cells are removed, and cells are filtered by mitochondrial content ($>5\%$) and extreme read counts ($5^{\text{th}}$--$95^{\text{th}}$ percentiles). We then apply stratified downsampling to preserve rare cell populations while controlling computational cost. Finally, log-normalized expression profiles are projected into a dense PCA feature space $\mathbf{X}^{pca} \in \mathbb{R}^{N \times 50}$.

\subsection{Statistical Alignment Edges}
Before introducing biological priors, DOGMA constructs statistical mutual-nearest-neighbor edges within each species. For species $s$, let $\mathcal{I}_s = \{i \mid s_i = s\}$ denote the index set of cells from species $s$. The alignment branch is
\begin{equation}
\mathcal{E}_{\text{Align}} =
\bigcup_{s \in \mathcal{S}}
\left\{
(i,j)
\;\middle|\;
i,j \in \mathcal{I}_s,\
j \in \mathrm{NN}^{\cos}_{k_{\text{mnn}}}(i;\mathbf{X}^{pca}),\
i \in \mathrm{NN}^{\cos}_{k_{\text{mnn}}}(j;\mathbf{X}^{pca})
\right\}.
\end{equation}
This branch preserves the local expression manifold before ontology or phylogeny priors are injected.

\subsection{Ontology-Guided Semantic Masking}
The ontology branch restricts candidate neighbors using cell-type semantics, but it is applied only to nodes whose cell-type labels are available for graph construction. Let
\begin{equation}
s^{pca}_{ij} = \cos(\mathbf{x}^{pca}_i,\mathbf{x}^{pca}_j)
\end{equation}
denote cosine similarity in PCA space, and let $m_i \in \{0,1\}$ indicate whether node $i$ is a training/reference cell with an available cell-type label. Validation, test, query, and otherwise unlabeled nodes are assigned $m_i=0$; their cell-type labels are never used by this branch. For Brain and Multi, we use Cell Ontology (CL) as the semantic reference $O_b$; for the Human benchmark, we use HCAO to better capture organ-specific cell-type granularity. The ontology distance threshold $\tau_b$ is selected per benchmark from $\{1,2\}$ and capped at $\tau_b \le 2$, because larger distances may connect biologically divergent cell types as neighbors, such as endothelial cells and erythroid lineage cells.
\begin{equation}
\resizebox{0.90\linewidth}{!}{$\displaystyle
\mathcal{E}_{\text{Onto}} =
\left\{
(i,j)
\middle|\!
j\in\operatorname{TopK}_{c\in\mathcal{C}^{\text{Onto}}_i}\!\left(s^{pca}_{ic},k_{\text{onto}}\right),\!
\mathcal{C}^{\text{Onto}}_i\!=\!
\left\{
c\in\mathcal{C}
\middle|\!
m_i=m_c=1,\ d_{O_b}(y_i,y_c)\le\tau_b
\right\}
\right\}.
$}
\end{equation}
Ontology defines biologically admissible candidates among labeled training/reference cells, while top-$k$ controls branch sparsity. Because $\mathcal{C}^{\text{Onto}}_i$ requires $m_i=m_c=1$, ontology-derived edges are never incident to validation/test/query nodes whose labels are unavailable. Such nodes are connected only through label-free graph branches, which prevents cell-type label leakage in annotation and transfer settings.

\subsection{Phylogenetically Stratified Cross-Species Edges}
For cross-species benchmarks, DOGMA constructs cross-species edges by first matching Leiden clusters in a shared-gene bridging space and then selecting edges from an expanded top-$K_{\mathrm{cand}}$ candidate pool within matched cluster pairs. Reciprocal $k_b$-NN pairs provide high-confidence anchors and a bidirectional scoring term, while directional top-$K_{\mathrm{cand}}$ neighbors broaden the candidate pool before final selection. The number of edges admitted for each species pair is governed by a divergence-time-aware budget:
\begin{equation}
\pi_{ab} = \exp\!\left(-\frac{t_{ab}}{\tau}\right),
\qquad
B_{ab} = \left\lfloor B \cdot
\frac{\pi_{ab}}{\sum_{(a',b')} \pi_{a'b'}} \right\rfloor,
\end{equation}
where $t_{ab}$ is the estimated divergence time for species pair $(a,b)$, $\tau$ is a temperature parameter, and $B$ is the total cross-species edge budget. Candidate edges are admitted in descending score order until $B_{ab}$ is reached or a per-node degree cap is met. The full specification of bridging-space construction, cluster matching, candidate scoring, and edge selection is given in \appref{app:phylo_details}. For the single-species Human benchmark, no phylogeny-derived branch is instantiated, so $\mathcal{E}_{\text{Phy}} = \varnothing$.

\subsection{Graph Assembly}
After constructing the branches independently, DOGMA merges them by set union and explicit symmetrization:
\begin{equation}
\mathcal{E}_{\text{raw}} =
\mathcal{E}_{\text{Align}} \cup
\mathcal{E}_{\text{Onto}} \cup
\mathcal{E}_{\text{Phy}},
\qquad
\mathcal{E} = \mathcal{E}_{\text{raw}}
\cup
\left\{ (j,i) \mid (i,j) \in \mathcal{E}_{\text{raw}} \right\},
\end{equation}
with binary adjacency
\begin{equation}
A_{ij} = \mathbb{I}\left[(i,j) \in \mathcal{E}\right].
\end{equation}
Set union is used because the three branches encode complementary relational axes: statistical proximity, semantic relatedness, and evolutionary conservation. Taking the intersection would discard edges supported by only one prior. The graph is kept unweighted so that attention-based backbones such as GAT can learn neighbor importance during training.

\subsection{GO-Based Semantic Feature Construction}
To inject functional semantics, DOGMA constructs a GO-based feature vector for each cell in a species-specific manner. For each species $s \in \mathcal{S}$, we select the top-200 highly variable genes from cells of that species and map them to GO terms via NCBI gene annotation databases (\texttt{gene\_info} and \texttt{gene2go}) using the species-specific taxonomy identifier. Candidate GO terms are ranked by a coverage-specificity score: terms receive higher priority when they annotate more selected HVGs for that species, while a specificity factor down-weights overly broad functional terms. We retain the top $D_{go}=200$ terms under this score; this default was selected after the fixed-hyperparameter sweep in \appref{app:go200_sensitivity}.

For cell $c_i$ of species $s$ and GO term $t$, let $\mathcal{G}_{s,t} = \{g \in G_s \mid t \in \mathrm{GO}(g)\}$ denote the set of selected HVGs annotated with term $t$. The raw functional score is the mean expression over annotated genes:
\begin{equation}
\tilde{z}_{i,t}
= \frac{1}{|\mathcal{G}_{s,t}|}
  \sum_{g \in \mathcal{G}_{s,t}} x^{raw}_{i,g}.
\end{equation}
The final knowledge feature is obtained by Z-score normalization across cells within each term:
\begin{equation}
z_{i,t}
= \frac{\tilde{z}_{i,t} - \mu_t}{\sigma_t},
\qquad
\mu_t = \frac{1}{|\mathcal{I}_s|}\sum_{j \in \mathcal{I}_s} \tilde{z}_{j,t},
\quad
\sigma_t = \mathrm{Std}\!\left(\{\tilde{z}_{j,t}\}_{j \in \mathcal{I}_s}\right).
\end{equation}
The final node representation concatenates the observation and knowledge views:
\begin{equation}
\mathbf{H}_i = [\mathbf{x}^{pca}_i \parallel \mathbf{z}_i].
\end{equation}

\section{Dataset Details}
\label{app:datasets}
To comprehensively evaluate the robustness and generalization capabilities of DOGMA, we curated three distinct graph benchmarks derived from the CELLxGENE database. These datasets differ in biological complexity, ranging from single-organ cross-species conservation to complex multi-organ and multi-species heterogeneity. The raw gene expression matrices were processed to retain the specific subsets listed below.

\subsection{Benchmark Construction}
\paragraph{1. Brain Benchmark (Level 1: Cross-Species, Single-Organ).}
This benchmark focuses on modeling evolutionary conservation within the brain cortex across primates and rodents. It integrates data from three major studies covering Chimpanzee, Marmoset and Mouse.
\begin{itemize}[leftmargin=1.5em]
\item \textbf{Source 1:} \textit{Evolution of cellular diversity in primary motor cortex of human, marmoset monkey, and mouse}. We utilized the Marmoset (non-neuron) subset containing 4,289 cells and 14,409 genes.
\item \textbf{Source 2:} \textit{Transcriptional profiling of murine oligodendrocyte precursor cells across the lifespan}. This subset includes 38,807 Mus musculus cells with 51,727 genes.
\item \textbf{Source 3:} \textit{Molecular and cellular evolution of the primate dorsolateral prefrontal cortex (dlPFC)}. We integrated two high-resolution subsets: Pan troglodytes (158,099 cells, 23,534 genes) and Callithrix jacchus (149,467 cells, 28,346 genes).
\end{itemize}

\paragraph{2. Human Benchmark (Level 2: Single-Species, Multi-Organ).}
Derived from the \textit{Tabula Sapiens} atlas, this benchmark evaluates the model's ability to capture tissue heterogeneity within a single species (Homo sapiens). It comprises three distinct organs with high-dimensional feature spaces.
\begin{itemize}[leftmargin=1.5em]
\item \textbf{Lung:} 65,847 cells, 61,759 genes.
\item \textbf{Small Intestine:} 42,036 cells, 61,759 genes.
\item \textbf{Tongue:} 38,754 cells, 61,759 genes.
\end{itemize}

\paragraph{3. Multi Benchmark (Level 3: Cross-Species, Multi-Organ).}
This is the most challenging scenario, designed to test robustness against simultaneous domain shifts in species and tissue types. It aggregates data from four studies.
\begin{itemize}[leftmargin=1.5em]
\item \textbf{Cortex (Marmoset):} Sourced from \textit{Comparative transcriptomics reveals human-specific cortical features} (75,861 cells, 12,897 genes).
\item \textbf{Cortex (Macaque):} Also sourced from the Great Apes study (89,136 cells, 19,784 genes).
\item \textbf{Thymus (Mouse):} Sourced from \textit{Single-cell multiomic analysis of thymocyte development} (29,408 cells, 15,942 genes).
\item \textbf{Blood (Human):} Homo sapiens blood subset with tissue ontology term UBERON:0000178, included in the evaluation file \texttt{graph3\_Homo\_sapiens.h5ad} after downsampling (1,998 cells).
\end{itemize}

\subsection{Supplementary Release}
\label{app:supplementary_release}
The anonymized supplementary ZIP accompanying this submission provides the code and configuration needed to reproduce the main DOGMA experiments. The release covers graph construction, downstream evaluation, and random-seed configuration. For upstream baseline repositories without an explicit redistribution license, the release records the upstream source and reproduction configuration but does not redistribute the upstream source code.

\subsection{Data and Asset Licenses}
\label{app:data_asset_licenses}
\tabref{tab:data_asset_licenses} summarizes the source studies, data portals, software assets, license or terms-of-use information, and their use in this paper. Dataset-specific access information and release versions are retained in the supplementary documentation.
The ontology source files \texttt{cl.owl}, \texttt{hcao.owl}, and \texttt{go.obo} were downloaded on 2026-04-12.

\begin{table}[H]
    \centering
    \caption{\textbf{Data and Asset Licenses.} Summary of existing data, ontology, annotation, and baseline assets used in this paper.}
    \label{tab:data_asset_licenses}
    \scriptsize
    \setlength{\tabcolsep}{3pt}
    \renewcommand{\arraystretch}{1.12}
    \begin{tabularx}{\textwidth}{p{0.16\textwidth}p{0.25\textwidth}p{0.31\textwidth}X}
        \toprule
        \textbf{Source study / asset} & \textbf{Dataset portal / asset source} & \textbf{License / terms} & \textbf{Use in this paper} \\
        \midrule
        CZ CELLxGENE Discover & \href{https://cellxgene.cziscience.com/}{CELLxGENE portal}; \href{https://cellxgene.cziscience.com/tos}{Terms of Service} & CELLxGENE Terms of Service; source-study and dataset-level terms retained; no re-identification attempted. & Access portal for public scRNA-seq matrices and metadata used to construct benchmark splits. \\
        Tabula Sapiens & \href{https://cellxgene.cziscience.com/collections/e5f58829-1a66-40b5-a624-9046778e74f5}{CELLxGENE collection}; \href{https://doi.org/10.1126/science.abl4896}{source study} & CELLxGENE terms plus original public source-study terms. & Source dataset for the Human benchmark. \\
        Gene Ontology (GO) & \href{https://geneontology.org/docs/go-citation-policy/}{GO citation and license page} & Creative Commons Attribution 4.0 (CC BY 4.0); release/version recorded in release documentation. & GO functional terms and semantic gene-feature augmentation. \\
        Cell Ontology (CL) & \href{https://obofoundry.org/ontology/cl.html}{OBO Foundry CL page} & Creative Commons Attribution 4.0 (CC BY 4.0). & Cell-type ontology edges for Brain and Multi benchmarks. \\
        Human Cell Atlas Ontology (HCAO) & \href{https://github.com/HumanCellAtlas/ontology}{HumanCellAtlas/ontology}; HCAO PURL & Public ontology source; no separate license file identified in the repository. Source/version are cited, and users are directed to the upstream source for original ontology files. & HCAO-derived organ-specific cell-type edges for the Human benchmark. \\
        NCBI Gene \texttt{gene\_info} and \texttt{gene2go} & \href{https://ftp.ncbi.nlm.nih.gov/gene/DATA/README}{NCBI Gene FTP README}; \href{https://www.ncbi.nlm.nih.gov/home/about/policies/}{NCBI policies} & NCBI molecular data usage policy; NCBI itself places no restrictions on molecular data use or distribution, subject to source-submitter rights caveats. & Gene-to-GO mapping for species-specific GO feature construction. \\
        scGPT & \href{https://github.com/bowang-lab/scGPT}{bowang-lab/scGPT} & MIT license. & Representation-centric transcriptome baseline. \\
        scCello & \href{https://github.com/DeepGraphLearning/scCello}{DeepGraphLearning/scCello} & README badge states MIT; no separate root license file identified. & Ontology-aware representation baseline. \\
        SATURN & \href{https://github.com/snap-stanford/SATURN}{snap-stanford/SATURN} & MIT license. & Cross-species alignment baseline. \\
        scMoGNN & \href{https://github.com/wehos/scmognn}{wehos/scmognn} & Public GitHub source; no explicit license file identified. Used only to reproduce baseline results; upstream source code is not redistributed. & Graph-structured baseline reproduced from the GitHub source. \\
        scPriorGraph & \href{https://github.com/ChrisOliver2345/scPriorGraph}{ChrisOliver2345/scPriorGraph} & Public GitHub source; no explicit license file identified. Used as a cited baseline or reimplementation; upstream source code is not redistributed. & Graph-structured prior baseline. \\
        KNN and MNN baselines & Local implementation following cited algorithms and package documentation & Algorithmic baselines implemented in our code; dependency licenses are recorded in the supplementary environment files. & Neighborhood and statistical-alignment baselines. \\
        \bottomrule
    \end{tabularx}
\end{table}

\section{Empirical Experiment Settings}

\subsection{Data Representation: Tokenization Strategies}
For the controlled architecture-complexity comparison in \figpartref{fig:teaser}{a}, we distinguish two input tokenization strategies: \textit{Cell Tokenization} and \textit{Gene Tokenization}.

\subsubsection{Cell Tokenization}
\begin{itemize}
    \item \textbf{Definition:} Each token represents a single cell $\mathbf{c}_i \in \mathbb{R}^{d_{gene}}$. A set of $N$ cells forms the input sequence/graph.
    \item \textbf{Setup ($N=200$):} Consistent with our proposed method, we sample a mini-batch of $N=200$ cells per iteration. This results in a $200 \times 200$ interaction matrix (Adjacency or Attention map) representing cell-cell similarity.
\end{itemize}

\subsubsection{Gene Tokenization}
\begin{itemize}
    \item \textbf{Definition:} Each token represents a specific gene $\mathbf{g}_j \in \mathbb{R}^{d_{cell}}$. The input represents the expression profile of a single cell across gene tokens.
    \item \textbf{Setup:}
    \begin{enumerate}
        \item \textbf{Gene GNN ($N=200$):} To match the GNN constraints, we select the top-200 Highly Variable Genes (HVGs) to construct a $200 \times 200$ gene regulatory graph.
        \item \textbf{Gene Transformer ($N=\mathrm{All}$):} Following standard scBERT-like implementations, this baseline processes the full sequence of expressed genes (approx.\ 2,000+), resulting in significant computational overhead.
    \end{enumerate}
\end{itemize}

\subsection{Model Architectures and Complexity Analysis}

We compare four distinct configurations corresponding to the data points in \figpartref{fig:teaser}{a}.

\paragraph{1. Cell Token GNN (Ours, $\star$ Red Star).}
\textbf{Analysis:} By leveraging the geometric prior through a dynamically learned adjacency matrix $\mathbf{A} \in \mathbb{R}^{200 \times 200}$, this model efficiently captures cell manifolds. With a compact dimension of 128 and shared graph weights, it achieves optimal accuracy with only 1.0M parameters and fast inference ($10^0$ scale). This controlled architecture proxy is used only for the teaser complexity comparison and is not the downstream DOGMA GCN/GAT evaluator, whose task-specific parameter counts are reported in \appref{app:dogma-gcn-gat-params}.

\paragraph{2. Cell Token Transformer (Baseline, \textcolor{blue}{$\bullet$} Blue Circle).}
\textbf{Analysis:} It requires larger hidden dimension (256) and FFN (ratio=4) to approximate relationships, tripling parameters (3.5M) while maintaining similar speed ($N=200$).

\paragraph{3. Gene Token GNN (Baseline, \textcolor{gray}{$\blacklozenge$} Gray Diamond).}
\textbf{Analysis:} This model constructs a $200 \times 200$ adjacency matrix representing Gene Regulatory Networks (GRNs). Since gene regulation logic is inherently more complex than cell similarity, we utilize a significantly wider network ($d=512$) to capture high-order interactions. This results in a larger model (3.2M parameters) and slightly slower inference ($10^1$ scale) compared to the Cell Token GNN.

\paragraph{4. Gene Token Transformer (Baseline, \textcolor{gray}{$\bullet$} Gray Circle).}
\textbf{Analysis:} It processes full sequences with $O(N^2)$ complexity (slow $10^2$ inference). Hidden dimension is constrained to $d=64$ to avoid OOM errors, yielding 0.5M parameters.

\subsection{Implementation Details}
All models were implemented using PyTorch and trained on a single NVIDIA RTX PRO 6000 GPU\@. The hyperparameter configurations were chosen to ensure the parameter counts match \figpartref{fig:teaser}{a}.

\section{Experimental Settings}
\label{sec:experimental_settings}

\subsection{Data Preprocessing}
We evaluate DOGMA on three comprehensive benchmarks constructed from the CELLxGENE database to assess performance across varying biological complexities. The Human Benchmark comprises lung, small intestine, and tongue tissues. Uniquely for this human-specific dataset, we adopt the Human Cell Atlas Ontology (HCAO) instead of the standard CL to provide more granular, organ-specific semantic guidance during topology construction. Because this benchmark is single-species, DOGMA does not instantiate a phylogeny-derived edge branch for Human; the final graph is constructed from MNN and HCAO-derived edges only. All datasets undergo stratified downsampling to balance class distributions and feature initialization via PCA ($d=50$) on log-normalized gene expression counts.

\subsection{Task Settings and Evaluation}
Our evaluation framework spans three distinct learning paradigms. For supervised classification tasks, including cell type, tissue, and development stage prediction, we employ a stratified split of 50\% training, 20\% validation, and 30\% testing. For strict zero-shot cell-type evaluation, we use a label-strict transductive protocol: all cells may remain in the graph for message passing, but strictness refers to label visibility, so unseen cell-type labels are withheld from ontology-edge construction, supervision, classifier outputs, and prototype computation. We first partition the cell-type label set into seen and unseen classes:
\begin{equation}
\mathcal{Y} = \mathcal{Y}_{\text{seen}} \sqcup \mathcal{Y}_{\text{unseen}},
\end{equation}
and optimize a two-layer graph encoder together with a linear classifier using only seen-class supervision:
\begin{equation}
\mathbf{h}_i = f_{\theta}(\mathbf{H}^{(0)}, \mathbf{A})_i,
\qquad
\hat{\mathbf{y}}^{\text{seen}}_i = \mathrm{Linear}(\mathbf{h}_i),
\qquad
\mathcal{L}_{\text{sup}} =
\sum_{i \in \mathcal{V}_{\text{seen}}}
\mathrm{CE}\!\left(\hat{\mathbf{y}}^{\text{seen}}_i, y_i\right),
\end{equation}
where the classifier output space only covers $\mathcal{Y}_{\text{seen}}$. Cells from unseen classes are not used to train the classifier or prototypes and do not contribute to the supervised loss. Unseen-class nodes may remain in the graph during message passing, but ontology-derived edges requiring unseen target labels are disabled; MNN and phylogeny edges remain available only when they do not require hidden target labels. Unless otherwise noted, we then compute seen-class prototypes
\begin{equation}
\mathbf{p}_c =
\frac{1}{|\mathcal{V}_c|}
\sum_{i \in \mathcal{V}_c} \mathbf{h}_i,
\qquad c \in \mathcal{Y}_{\text{seen}},
\qquad
\hat{c}_i =
\arg\max_{c \in \mathcal{Y}_{\text{seen}}}
\cos(\mathbf{h}_i, \mathbf{p}_c),
\qquad i \in \mathcal{V}_{\text{unseen}}.
\end{equation}
Final zero-shot predictions are therefore not taken from the classifier head directly. The nearest seen-class prototype index $\hat{c}_i$ is treated as a cluster assignment over unseen cells, and we compare these assignments with the true unseen cell-type labels using ARI; NMI/AMI are used in the same cluster-label comparison where reported. For unsupervised clustering, no GCN/GAT encoder is trained. We run Leiden community detection directly on the constructed cell graph $\mathbf{A}$, whose topology uses MNN plus CL and phylogeny for Brain/Multi, and MNN plus HCAO for Human. We report Adjusted Rand Index (ARI) and Adjusted Mutual Information (AMI) to quantify the alignment between derived communities and ground-truth annotations. GO-derived node features are therefore not consumed by this fixed-graph clustering evaluator; removing GO features leaves the clustering result unchanged when the graph topology is held fixed.

\subsection{Uncertainty Reporting}
\label{app:uncertainty_reporting}
Unless otherwise stated, repeated experimental results report the mean and 95\% confidence-interval half-width over random seeds. For a metric value $x_i$ from seed $i$, $i=1,\dots,n$, we compute
\begin{equation}
\bar{x} = \frac{1}{n}\sum_{i=1}^{n} x_i,
\qquad
s = \sqrt{\frac{1}{n-1}\sum_{i=1}^{n}(x_i-\bar{x})^2},
\qquad
\mathrm{SEM} = \frac{s}{\sqrt{n}}.
\end{equation}
The reported 95\% CI half-width uses a normal approximation:
\begin{equation}
\mathrm{CI}_{95} = 1.96 \cdot \mathrm{SEM},
\qquad
\text{reported value} = \bar{x} \pm \mathrm{CI}_{95}.
\end{equation}
For the main supervised classification benchmark, results are computed over random seeds, typically ten seeds from 42 to 51; for $n=10$, the half-width is $1.96 \cdot s / \sqrt{10}$.

\paragraph{SATURN strict zero-shot readout diagnostic.}
For SATURN on Brain and Human, metadata alignment was complete (Brain: 5,771 cells with zero missing cell-type labels; Human: 6,030 cells with zero missing cell-type labels). The main comparison uses an MLP adapter followed by nearest seen-class prototypes, yielding SATURN ARI of 0.2511 (95\% CI: [0.2200, 0.2823], $n=10$) on Brain and 0.5635 (95\% CI: [0.5574, 0.5695], $n=10$) on Human. This readout is fragile on the Brain split, which contains only 145 unseen cells distributed across 11 rare classes, including classes represented by 2, 2, and 1 cells. Under the same SATURN final embedding and Brain split, ARI changes from 0.250 with MLP-latent nearest seen prototypes to 0.471 with MLP-latent unseen KMeans, 0.504 with raw-embedding unseen KMeans, and 0.5254 with final all-standardized unseen KMeans (95\% CI: [0.5174, 0.5334], $n=10$). The Human split is more stable (0.562 with MLP-latent nearest seen prototypes, 0.540 with MLP-latent unseen KMeans, and 0.5728 with final raw-embedding unseen KMeans; 95\% CI: [0.5521, 0.5935], $n=10$). We treat unseen KMeans as a diagnostic rather than a main strict zero-shot score because it uses $n_{\mathrm{clusters}}$ equal to the true number of unseen cell types, which introduces oracle information unavailable to DOGMA and the prototype-based baselines.

\subsection{Downstream Resource Usage}
\label{app:downstream_resource_usage}

We report runtime and GPU memory for the graph3 100\% prepared-forward downstream evaluation window ($n=2$). Memory columns use CUDA/PyTorch allocated and reserved counters from that window; they are not process-level peak GPU memory and are not averaged across all benchmark graphs. This measurement excludes graph construction, hyperparameter search, model downloading, and warm-up overheads, and therefore should not be interpreted as end-to-end training or graph-construction cost. DOGMA aggregates the GCN and GAT variants by averaging their inference time and reporting the union of their observed memory ranges; the reserved-memory average used for ratio calculations is 146 MB.

\begin{table}[H]
    \centering
    \caption{\textbf{Downstream Resource Usage.} Runtime and GPU memory during downstream evaluation. SATURN is omitted because raw comparable SATURN forward timing was unavailable.}
    \label{tab:downstream_resource_usage}
    \scriptsize
    \setlength{\tabcolsep}{5pt}
    \begin{tabular}{lccc}
        \toprule
        \textbf{Method} & \textbf{Inference Time (s)} & \textbf{Allocated Memory (MB)} & \textbf{Reserved Memory (MB)} \\
        \midrule
        DOGMA & 0.0013 & 70--73 & 110--182 \\
        scPriorGraph & 0.0139 & 352 & 522 \\
        scMoGNN & 0.0872 & 5,851 & 6,323 \\
        scGPT & 3.9213 & 784 & 856 \\
        scCello & 7.8761 & 5,446 & 6,324 \\
        \bottomrule
    \end{tabular}
\end{table}

\subsection{End-to-End Pipeline Cost}
\label{app:e2e_pipeline_cost}

To complement the downstream measurements, we also report the end-to-end wall-clock and resource footprint for the full DOGMA pipeline, including graph construction, hyperparameter search, and final repeated evaluation. The measured pipeline took 8,949.4 seconds (2h 29m 9.4s) in total. Most of the time was spent on the three downstream hyperparameter searches, while graph construction and final 10-seed evaluation were comparatively lightweight.

\begin{table}[H]
    \centering
    \caption{\textbf{End-to-End DOGMA Pipeline Cost.} Wall-clock time, peak resources, and final artifact sizes for the full graph construction and downstream search pipeline.}
    \label{tab:e2e_pipeline_cost}
    \scriptsize
    \setlength{\tabcolsep}{6pt}
    \begin{tabular}{llr}
        \toprule
        \textbf{Category} & \textbf{Item} & \textbf{Value} \\
        \midrule
        Total wall-clock & Full pipeline & 8,949.4 s (2h 29m 9.4s) \\
        \midrule
        Runtime breakdown & Graph construction & 158.2 s \\
        Runtime breakdown & Three-task search & 8,790.6 s \\
        Runtime breakdown & Final 10-seed evaluation & 0.54 s \\
        \midrule
        Search breakdown & Classification, 500 trials & 3,188.0 s \\
        Search breakdown & Strict zero-shot, 500 trials & 3,709.9 s \\
        Search breakdown & Clustering, 500 trials & 1,892.7 s \\
        \midrule
        Peak resource & GPU memory & $\sim$746 MB \\
        Peak resource & RAM & $\sim$1,266 MB \\
        \midrule
        Artifact size & Three \texttt{best\_cell\_graph.pt} files & 32.47 MB \\
        Artifact size & Classification graph & 9.97 MB \\
        Artifact size & Strict zero-shot graph & 11.45 MB \\
        Artifact size & Clustering graph & 11.04 MB \\
        \bottomrule
    \end{tabular}
\end{table}

\subsection{DOGMA GCN/GAT Model Parameters}
\label{app:dogma-gcn-gat-params}

\tabref{tab:dogma-gcn-gat-hparams} summarizes the GCN/GAT training configuration used by DOGMA\@. The supervised node-classification evaluator uses a three-layer GCN/GAT classifier for each metadata prediction task. The strict zero-shot evaluator uses a two-layer GCN/GAT feature extractor and trains a separate linear classifier on the seen cell-type classes before prototype-based evaluation on unseen cells. The clustering evaluator is fixed-graph Leiden on the constructed adjacency matrix, rather than a GCN/GAT embedding pipeline, and therefore does not train a GCN or GAT\@.

\begin{table}[H]
    \centering
    \caption{\textbf{DOGMA GCN/GAT Hyperparameters.}}
    \label{tab:dogma-gcn-gat-hparams}
    \scriptsize
    \setlength{\tabcolsep}{5pt}
    \begin{tabular}{lll}
        \toprule
        \textbf{Parameter} & \textbf{Supervised classification} & \textbf{Zero-shot feature extractor} \\
        \midrule
        Input node feature dimension & 250 & 250 \\
        GCN architecture & 3 $\times$ GCNConv & 2 $\times$ GCNConv \\
        GAT architecture & 3 $\times$ GATConv & 2 $\times$ GATConv \\
        Hidden dimension & 128 & 64 \\
        Output embedding dimension & N/A & 64 \\
        GAT heads & 4 & 4 in layer 1; 1 in layer 2 \\
        Per-head channels & 32 & 16 in layer 1; 64 in layer 2 \\
        Dropout & 0.5 & 0.2 \\
        Epochs & 100 & 100 \\
        Optimizer & Adam & Adam \\
        Learning rate & 0.01 & 0.01 \\
        Weight decay & $5 \times 10^{-4}$ & $5 \times 10^{-4}$ for extractor \\
        Auxiliary classifier & None & Linear(64, seen classes), Adam, lr 0.01 \\
        Evaluation scope & Cell type, stage, tissue, sex & Cell type only \\
        \bottomrule
    \end{tabular}
\end{table}

Parameter counts use trainable PyTorch/PyG parameters (\texttt{requires\_grad}), including BatchNorm and classifier heads but excluding optimizer state and graph data.

\begin{table}[H]
    \centering
    \caption{\textbf{DOGMA Supervised Node-Classification Parameter Counts.}}
    \label{tab:dogma-classification-param-counts}
    \scriptsize
    \setlength{\tabcolsep}{5pt}
    \begin{tabular}{llrrr}
        \toprule
        \textbf{Graph} & \textbf{Task} & \textbf{Classes} & \textbf{GCN params} & \textbf{GAT params} \\
        \midrule
        graph1 & Cell type & 26 & 69,274 & 70,042 \\
        graph1 & Development stage & 6 & 66,694 & 67,462 \\
        graph1 & Tissue & 3 & 66,307 & 67,075 \\
        graph1 & Sex & 2 & 66,178 & 66,946 \\
        graph2 & Cell type & 61 & 73,789 & 74,557 \\
        graph2 & Development stage & 7 & 66,823 & 67,591 \\
        graph2 & Tissue & 8 & 66,952 & 67,720 \\
        graph2 & Sex & 2 & 66,178 & 66,946 \\
        graph3 & Cell type & 43 & 71,467 & 72,235 \\
        graph3 & Development stage & 14 & 67,726 & 68,494 \\
        graph3 & Tissue & 3 & 66,307 & 67,075 \\
        graph3 & Sex & 2 & 66,178 & 66,946 \\
        \bottomrule
    \end{tabular}
\end{table}

\begin{table}[H]
    \centering
    \caption{\textbf{DOGMA Strict Zero-Shot Cell-Type Evaluation Parameter Counts.} The total includes the feature extractor plus the auxiliary linear classifier over seen classes.}
    \label{tab:dogma-zeroshot-param-counts}
    \scriptsize
    \setlength{\tabcolsep}{4pt}
    \begin{tabular}{lrrrrrr}
        \toprule
        \textbf{Graph} & \textbf{Seen} & \textbf{Unseen} & \textbf{GCN extractor} & \textbf{GAT extractor} & \textbf{Aux.\ classifier} & \textbf{GCN/GAT total} \\
        \midrule
        graph1 & 15 & 11 & 20,224 & 20,480 & 975 & 21,199 / 21,455 \\
        graph2 & 36 & 25 & 20,224 & 20,480 & 2,340 & 22,564 / 22,820 \\
        graph3 & 25 & 18 & 20,224 & 20,480 & 1,625 & 21,849 / 22,105 \\
        \bottomrule
    \end{tabular}
\end{table}

\section{Algorithm Overview}
\label{app:algorithm}

\begin{algorithm}[H]
\fontsize{7}{6.7}\selectfont
\setlength{\parskip}{0pt}
\caption{DOGMA: Data-centric Ontology-Guided Modeling Approach}
\label{alg:dogma}
\begin{algorithmic}[1]
\Require $\mathbf{X}^{raw}$; cell-type labels $\{y_i\}$; species labels $\{s_i\}$; Cell Ontology $\mathcal{G}_{CL}$; Phylogenetic tree $\mathcal{T}_{phy}$ with divergence times $\{t_{ab}\}$; Gene Ontology $\mathcal{G}_{GO}$; hyperparameters for MNN, ontology masking, cross-species bridging, and GO fusion
\Ensure Adjacency matrix $\mathbf{A} \in \{0,1\}^{N \times N}$; node feature matrix $\mathbf{H} \in \mathbb{R}^{N \times (50 + D_{go})}$
\Statex
\Statex \textbf{--- Phase 1: Data Curation and Feature Initialization ---}
\State Filter genes expressed in $< 3$ cells
\State Filter cells with mitochondrial content $> 5\%$ or read counts outside the $[5^{\text{th}}, 95^{\text{th}}]$ percentile range
\State Perform stratified downsampling to balance cell-type distributions
\State $\mathbf{X}^{pca} \gets \mathrm{PCA}(\mathrm{LogNorm}(\mathbf{X}^{raw}),\, d{=}50)$
\Statex
\Statex \textbf{--- Phase 2: Multi-Branch Topology Construction ---}
\Statex \textit{Branch 1: Statistical Alignment (MNN)}
\For{each species $s \in \mathcal{S}$}
    \State $\mathcal{I}_s \gets \{i \mid s_i=s\}$
    \For{each cell $i \in \mathcal{I}_s$}
        \State $\mathcal{N}_i \gets k_{\text{mnn}}\text{-nearest neighbors of } i \text{ within } \mathcal{I}_s \text{ in } \mathbf{X}^{pca} \text{ (cosine)}$
    \EndFor
\EndFor
\State $\mathcal{E}_{\text{Align}} \gets \{(i,j) \mid j \in \mathcal{N}_i \wedge i \in \mathcal{N}_j\}$ \Comment{Mutual nearest neighbors}
\Statex
\Statex \textit{Branch 2: Ontology-Guided Semantic Masking}
\State Select ontology $O_b$ and threshold $\tau_b$ based on benchmark $b$
\State $\mathcal{E}_{\text{Onto}} \gets \varnothing$
\For{each cell $i$ where label mask $m_i = 1$}
    \State $\mathcal{C}^{\text{Onto}}_i \gets \{c \in \mathcal{C} \mid m_c = 1 \wedge d_{O_b}(y_i, y_c) \le \tau_b\}$ \Comment{Ontology-admissible candidates}
    \State $\mathcal{E}_{\text{Onto}} \gets \mathcal{E}_{\text{Onto}} \cup \{(i,j) \mid j \in \operatorname{TopK}_{c \in \mathcal{C}^{\text{Onto}}_i}(s^{pca}_{ic},\, k_{\text{onto}})\}$
\EndFor
\Statex
\Statex \textit{Branch 3: Phylogenetic Stratification (cross-species only)}
\State $\mathcal{E}_{\text{Phy}} \gets \varnothing$
\If{benchmark involves multiple species}
    \State $B \gets \mathrm{round}(N\rho/2)$; \quad $Z_{\pi} \gets \sum_{(a,b)} \exp(-t_{ab}/\tau)$
    \For{each unordered species pair $(a,b)$, $a \ne b$}
        \State $\mathcal{G}_{ab} \gets$ standardized shared gene symbols between species $a$ and $b$
        \If{$|\mathcal{G}_{ab}| < G_{\min}$} \State \textbf{continue} \EndIf
        \State $\mathbf{Z}^{ab} \gets \mathrm{BridgeSpace}(\mathbf{X}^{raw}, \mathcal{G}_{ab}, G_{\max})$ \Comment{log-normalized shared genes}
        \State $\mathcal{U}_a,\mathcal{U}_b \gets \mathrm{Leiden}(\mathbf{Z}^{ab}_{\mathcal{C}_a}), \mathrm{Leiden}(\mathbf{Z}^{ab}_{\mathcal{C}_b})$
        \State Score cluster pairs by centroid cosine similarity and marker-gene overlap
        \State $\mathcal{P}_{ab} \gets$ mutually top-$K_c$ cluster pairs with score $\ge\theta$
        \State $\mathcal{Q}_{ab} \gets \varnothing$
        \For{each retained cluster pair $(u,v) \in \mathcal{P}_{ab}$}
            \State Generate reciprocal $k_b$-NN anchors and directional top-$K_{\mathrm{cand}}$ candidates in $\mathbf{Z}^{ab}$
            \State Add their union to $\mathcal{Q}_{ab}$; score reciprocal anchors by averaged bidirectional cosine and expanded candidates by max directional cosine
        \EndFor
        \State $B_{ab} \gets \left\lfloor B \cdot \exp(-t_{ab}/\tau) / Z_{\pi} \right\rfloor$
        \State Greedily admit top-scoring $\mathcal{Q}_{ab}$ edges subject to $B_{ab}$ and degree cap $d_{\max}$
    \EndFor
\EndIf
\Statex
\Statex \textbf{--- Phase 3: Graph Assembly ---}
\State $\mathcal{E}_{\text{raw}} \gets \mathcal{E}_{\text{Align}} \cup \mathcal{E}_{\text{Onto}} \cup \mathcal{E}_{\text{Phy}}$
\State $\mathcal{E} \gets \mathcal{E}_{\text{raw}} \cup \{(j,i) \mid (i,j) \in \mathcal{E}_{\text{raw}}\}$ \Comment{Symmetrization}
\State $A_{ij} \gets \mathbb{I}[(i,j) \in \mathcal{E}]$ \Comment{Unweighted binary adjacency}
\Statex
\Statex \textbf{--- Phase 4: Species-Specific Dual-View Feature Fusion ---}
\For{each species $s \in \mathcal{S}$}
    \State $\mathcal{I}_s \gets \{i \mid s_i = s\}$
    \State $G_s \gets \mathrm{SelectHVG}(\mathbf{X}^{raw}[\mathcal{I}_s],\, \text{top}{=}200)$ \Comment{Species-specific HVGs}
    \State $\textit{taxid}_s \gets \mathrm{SpeciesToTaxID}(s)$
    \For{each gene $g \in G_s$}
        \State $\mathrm{GO}(g) \gets \mathrm{NCBILookup}(\text{symbol}{=}g,\, \text{taxid}{=}\textit{taxid}_s)$ \Comment{gene\_info $+$ gene2go}
    \EndFor
    \State $\mathcal{T}_s \gets \mathrm{SelectGOTerms}(\{\mathrm{GO}(g) \mid g \in G_s\},\, D_{go})$ \Comment{Top-$D_{go}$ coverage-specificity terms; main setting uses $D_{go}=200$ (\appref{app:go200_sensitivity})}
    \For{each cell index $i \in \mathcal{I}_s$}
        \For{each GO term $t \in \mathcal{T}_s$}
            \State $\tilde{z}_{i,t} \gets \mathrm{Mean}(\{x^{raw}_{i,g} \mid g \in G_s \wedge t \in \mathrm{GO}(g)\})$ \Comment{Mean expression over annotated genes}
        \EndFor
    \EndFor
\EndFor
\State $\mathbf{z}_i \gets \text{Z-score normalize } (\tilde{z}_{i,1}, \dots, \tilde{z}_{i,D_{go}}) \text{ across cells per term}$
\State $\mathbf{H}_i \gets [\mathbf{x}^{pca}_i \parallel \mathbf{z}_i]$ \Comment{Concatenate statistical and semantic views}
\Statex
\State \Return $(\mathbf{A},\, \mathbf{H})$
\end{algorithmic}
\end{algorithm}

\section{Cross-Species Edge Construction Details}
\label{app:phylo_details}

This section provides the complete mathematical specification of the phylogenetically stratified cross-species edge construction described in Section~\ref{subsec:topology}.

\subsection{Bridging Feature Space}

For a species pair $(a,b)$, let $\mathcal{G}_{ab}$ denote the set of shared gene symbols after symbol standardization. If $|\mathcal{G}_{ab}| < G_{\min}$, no cross-species edges are constructed for this pair. For each cell $i$ and shared gene $g \in \mathcal{G}_{ab}$, we apply library-size normalization and log transformation:
\begin{equation}
\tilde{x}_{ig} = \log\!\left(1 + \frac{10^4\, x_{ig}}{\sum_{g'} x_{ig'}}\right).
\end{equation}
If $|\mathcal{G}_{ab}| > G_{\max}$, we select the top-$G_{\max}$ genes ranked by total variance across both species:
\begin{equation}
r_g = \mathrm{Var}\!\left(\tilde{X}^{(a)}_{\cdot g}\right) + \mathrm{Var}\!\left(\tilde{X}^{(b)}_{\cdot g}\right).
\end{equation}
Finally, per-species standardization is applied gene-wise:
\begin{equation}
z_{ig} = \frac{\tilde{x}_{ig} - \mu_g}{\sigma_g}.
\end{equation}

\subsection{Cluster-Level Matching}

Within the bridging space, we perform Leiden clustering on each species independently. Let $u$ and $v$ denote clusters from species $a$ and $b$ respectively, with centroid vectors:
\begin{equation}
\boldsymbol{\mu}_u = \frac{1}{|u|}\sum_{i \in u} \mathbf{z}_i,
\qquad
\boldsymbol{\mu}_v = \frac{1}{|v|}\sum_{j \in v} \mathbf{z}_j.
\end{equation}
For each cluster, let $M_u$ (resp.\ $M_v$) be the set of top-$N_m$ genes with highest centroid expression. The cluster matching score combines expression similarity and marker overlap:
\begin{equation}
s^{\mathrm{expr}}_{uv} = \cos(\boldsymbol{\mu}_u, \boldsymbol{\mu}_v),
\qquad
s^{\mathrm{mark}}_{uv} = \frac{|M_u \cap M_v|}{\min(|M_u|, |M_v|)},
\end{equation}
\begin{equation}
s_{uv} = \alpha\,\max(0,\, s^{\mathrm{expr}}_{uv}) + (1-\alpha)\, s^{\mathrm{mark}}_{uv},
\end{equation}
where $\alpha$ is the expression similarity weight. A cluster pair $(u,v)$ is retained if and only if $u$ and $v$ are mutually within each other's top-$K_c$ matches and $s_{uv} \ge \theta$.

\subsection{Cell-Level Candidate Edges}

Within each retained cluster pair $(u,v)$, DOGMA forms an expanded cell-level candidate pool in the bridging space. Reciprocal $k_b$-nearest-neighbor pairs serve as high-confidence anchors and define one scoring component, but reciprocity is not required for every final candidate edge. The arrows below indicate search direction only; the underlying metric is standard cosine similarity in the shared-gene bridge space. If cells $i \in u$ and $j \in v$ are mutual bridging neighbors, their anchor score is:
\begin{equation}
q^{\mathrm{mut}}_{ij} = s_{uv} \cdot \frac{\left(\sigma_{a \to b}(i,j) + \sigma_{b \to a}(j,i)\right)}{2},
\end{equation}
where $\sigma_{a \to b}(i,j)$ and $\sigma_{b \to a}(j,i)$ denote cosine similarities obtained from the $a \to b$ and $b \to a$ searches, respectively. DOGMA then adds expanded directional candidates: a pair is included if $j$ is among the top-$K_{\mathrm{cand}}$ neighbors of $i$ in the $a \to b$ search or $i$ is among the top-$K_{\mathrm{cand}}$ neighbors of $j$ in the $b \to a$ search. Expanded candidate edges are generated with score:
\begin{equation}
q^{\mathrm{exp}}_{ij} = s_{uv} \cdot \max\!\left(\sigma_{a \to b}(i,j),\, \sigma_{b \to a}(j,i)\right).
\end{equation}
If a cell pair appears in both sets, the maximum score is retained:
\begin{equation}
q_{ij} = \max\!\left(q^{\mathrm{mut}}_{ij},\, q^{\mathrm{exp}}_{ij}\right).
\end{equation}
Thus, reciprocal MNN is a high-confidence candidate and scoring signal, not a hard final filter. The final cross-species edge set is selected from the union of reciprocal anchors and expanded top-$K_{\mathrm{cand}}$ directional candidates.

\subsection{Budget Allocation and Edge Selection}

All candidate edges are ranked by $q_{ij}$ in descending order; no additional reciprocal-neighbor check is applied after candidate expansion. The total cross-species edge budget is:
\begin{equation}
B = \mathrm{round}\!\left(\frac{N \cdot \rho}{2}\right),
\end{equation}
where $N$ is the total number of cells and $\rho$ is the per-node cross-species edge density parameter. The budget for species pair $(a,b)$ is distributed proportionally to the divergence-time prior:
\begin{equation}
\pi_{ab} = \exp\!\left(-\frac{t_{ab}}{\tau}\right),
\qquad
B_{ab} = \left\lfloor B \cdot \frac{\pi_{ab}}{\sum_{(a',b')} \pi_{a'b'}} \right\rfloor.
\end{equation}
Candidate edges for pair $(a,b)$ are greedily admitted in descending $q_{ij}$ order subject to the constraint $d_i^{\mathrm{cross}} < d_{\max}$, where $d_i^{\mathrm{cross}}$ is the current cross-species degree of cell $i$ and $d_{\max}$ is a global degree cap. Selection terminates when $B_{ab}$ edges have been admitted or candidates are exhausted. Note that $B_{ab}$ is an upper bound: when valid candidates are insufficient or degree constraints are binding, fewer edges may be admitted, and unused budget is not redistributed.

\newpage
\input{checklist.tex}

\end{document}

%% file: checklist.tex
% !TEX root = ./neurips_2026.tex
\section*{NeurIPS Paper Checklist}

\begin{enumerate}

\item {\bf Claims}
    \item[] Question: Do the main claims made in the abstract and introduction accurately reflect the paper's contributions and scope?
    \item[] Answer: \answerYes{}.
    \item[] Justification: The abstract and Introduction state the paper's scoped contributions: a prior-guided graph construction pipeline using Cell Ontology, phylogeny, and Gene Ontology, plus empirical evaluation on multi-species and multi-organ benchmarks. The experimental sections report both strengths and exceptions, such as scGPT outperforming DOGMA on Brain cell-type annotation.
    \item[] Guidelines:
    \begin{itemize}
        \item The answer \answerNA{} means that the abstract and introduction do not include the claims made in the paper.
        \item The abstract and/or introduction should clearly state the claims made, including the contributions made in the paper and important assumptions and limitations. A \answerNo{} or \answerNA{} answer to this question will not be perceived well by the reviewers. 
        \item The claims made should match theoretical and experimental results, and reflect how much the results can be expected to generalize to other settings. 
        \item It is fine to include aspirational goals as motivation as long as it is clear that these goals are not attained by the paper. 
    \end{itemize}

\item {\bf Limitations}
    \item[] Question: Does the paper discuss the limitations of the work performed by the authors?
    \item[] Answer: \answerYes{}.
    \item[] Justification: Appendix~\ref{app:limitations} discusses limitations related to dependence on ontology and annotation quality, incompleteness of cross-species biological priors, and the finite coverage of the selected public benchmarks, tasks, and baselines.
    \item[] Guidelines:
    \begin{itemize}
        \item The answer \answerNA{} means that the paper has no limitation while the answer \answerNo{} means that the paper has limitations, but those are not discussed in the paper. 
        \item The authors are encouraged to create a separate ``Limitations'' section in their paper.
        \item The paper should point out any strong assumptions and how robust the results are to violations of these assumptions (e.g., independence assumptions, noiseless settings, model well-specification, asymptotic approximations only holding locally). The authors should reflect on how these assumptions might be violated in practice and what the implications would be.
        \item The authors should reflect on the scope of the claims made, e.g., if the approach was only tested on a few datasets or with a few runs. In general, empirical results often depend on implicit assumptions, which should be articulated.
        \item The authors should reflect on the factors that influence the performance of the approach. For example, a facial recognition algorithm may perform poorly when image resolution is low or images are taken in low lighting. Or a speech-to-text system might not be used reliably to provide closed captions for online lectures because it fails to handle technical jargon.
        \item The authors should discuss the computational efficiency of the proposed algorithms and how they scale with dataset size.
        \item If applicable, the authors should discuss possible limitations of their approach to address problems of privacy and fairness.
        \item While the authors might fear that complete honesty about limitations might be used by reviewers as grounds for rejection, a worse outcome might be that reviewers discover limitations that aren't acknowledged in the paper. The authors should use their best judgment and recognize that individual actions in favor of transparency play an important role in developing norms that preserve the integrity of the community. Reviewers will be specifically instructed to not penalize honesty concerning limitations.
    \end{itemize}

\item {\bf Theory assumptions and proofs}
    \item[] Question: For each theoretical result, does the paper provide the full set of assumptions and a complete (and correct) proof?
    \item[] Answer: \answerNA{}.
    \item[] Justification: The paper does not present theoretical results or theorem-proof claims; it provides methodological definitions, algorithmic specifications, and empirical evaluation.
    \item[] Guidelines:
    \begin{itemize}
        \item The answer \answerNA{} means that the paper does not include theoretical results. 
        \item All the theorems, formulas, and proofs in the paper should be numbered and cross-referenced.
        \item All assumptions should be clearly stated or referenced in the statement of any theorems.
        \item The proofs can either appear in the main paper or the supplemental material, but if they appear in the supplemental material, the authors are encouraged to provide a short proof sketch to provide intuition. 
        \item Inversely, any informal proof provided in the core of the paper should be complemented by formal proofs provided in appendix or supplemental material.
        \item Theorems and Lemmas that the proof relies upon should be properly referenced. 
    \end{itemize}

    \item {\bf Experimental result reproducibility}
    \item[] Question: Does the paper fully disclose all the information needed to reproduce the main experimental results of the paper to the extent that it affects the main claims and/or conclusions of the paper (regardless of whether the code and data are provided or not)?
    \item[] Answer: \answerYes{}.
    \item[] Justification: Sections~\ref{sec:methods} and~\ref{sec:experimental_settings} and Appendices~\ref{app:method_details},~\ref{app:datasets}, and~\ref{app:algorithm} describe preprocessing, graph construction, data sources, task splits, evaluation protocols, hyperparameters, and the algorithmic pipeline used for the main DOGMA results.
    \item[] Guidelines:
    \begin{itemize}
        \item The answer \answerNA{} means that the paper does not include experiments.
        \item If the paper includes experiments, a \answerNo{} answer to this question will not be perceived well by the reviewers: Making the paper reproducible is important, regardless of whether the code and data are provided or not.
        \item If the contribution is a dataset and\slash or model, the authors should describe the steps taken to make their results reproducible or verifiable. 
        \item Depending on the contribution, reproducibility can be accomplished in various ways. For example, if the contribution is a novel architecture, describing the architecture fully might suffice, or if the contribution is a specific model and empirical evaluation, it may be necessary to either make it possible for others to replicate the model with the same dataset, or provide access to the model. In general, releasing code and data is often one good way to accomplish this, but reproducibility can also be provided via detailed instructions for how to replicate the results, access to a hosted model (e.g., in the case of a large language model), releasing of a model checkpoint, or other means that are appropriate to the research performed.
        \item While NeurIPS does not require releasing code, the conference does require all submissions to provide some reasonable avenue for reproducibility, which may depend on the nature of the contribution. For example
        \begin{enumerate}
            \item If the contribution is primarily a new algorithm, the paper should make it clear how to reproduce that algorithm.
            \item If the contribution is primarily a new model architecture, the paper should describe the architecture clearly and fully.
            \item If the contribution is a new model (e.g., a large language model), then there should either be a way to access this model for reproducing the results or a way to reproduce the model (e.g., with an open-source dataset or instructions for how to construct the dataset).
            \item We recognize that reproducibility may be tricky in some cases, in which case authors are welcome to describe the particular way they provide for reproducibility. In the case of closed-source models, it may be that access to the model is limited in some way (e.g., to registered users), but it should be possible for other researchers to have some path to reproducing or verifying the results.
        \end{enumerate}
    \end{itemize}

\item {\bf Open access to data and code}
    \item[] Question: Does the paper provide open access to the data and code, with sufficient instructions to faithfully reproduce the main experimental results, as described in supplemental material?
    \item[] Answer: \answerYes{}.
    \item[] Justification: Appendix~\ref{app:supplementary_release} describes the anonymized supplementary ZIP, which provides code, graph-construction configuration, downstream evaluation configuration, and random-seed configuration for reproducing the main DOGMA experiments.
    \item[] Guidelines:
    \begin{itemize}
        \item The answer \answerNA{} means that paper does not include experiments requiring code.
        \item Please see the NeurIPS code and data submission guidelines (\url{https://neurips.cc/public/guides/CodeSubmissionPolicy}) for more details.
        \item While we encourage the release of code and data, we understand that this might not be possible, so \answerNo{} is an acceptable answer. Papers cannot be rejected simply for not including code, unless this is central to the contribution (e.g., for a new open-source benchmark).
        \item The instructions should contain the exact command and environment needed to run to reproduce the results. See the NeurIPS code and data submission guidelines (\url{https://neurips.cc/public/guides/CodeSubmissionPolicy}) for more details.
        \item The authors should provide instructions on data access and preparation, including how to access the raw data, preprocessed data, intermediate data, and generated data, etc.
        \item The authors should provide scripts to reproduce all experimental results for the new proposed method and baselines. If only a subset of experiments are reproducible, they should state which ones are omitted from the script and why.
        \item At submission time, to preserve anonymity, the authors should release anonymized versions (if applicable).
        \item Providing as much information as possible in supplemental material (appended to the paper) is recommended, but including URLs to data and code is permitted.
    \end{itemize}

\item {\bf Experimental setting/details}
    \item[] Question: Does the paper specify all the training and test details (e.g., data splits, hyperparameters, how they were chosen, type of optimizer) necessary to understand the results?
    \item[] Answer: \answerYes{}.
    \item[] Justification: Appendix~\ref{sec:experimental_settings} specifies preprocessing, train/validation/test splits, held-out transfer protocol, clustering protocol, optimizers, learning rates, epochs, architectures, and DOGMA GCN/GAT parameter counts.
    \item[] Guidelines:
    \begin{itemize}
        \item The answer \answerNA{} means that the paper does not include experiments.
        \item The experimental setting should be presented in the core of the paper to a level of detail that is necessary to appreciate the results and make sense of them.
        \item The full details can be provided either with the code, in appendix, or as supplemental material.
    \end{itemize}

\item {\bf Experiment statistical significance}
    \item[] Question: Does the paper report error bars suitably and correctly defined or other appropriate information about the statistical significance of the experiments?
    \item[] Answer: \answerYes{}.
    \item[] Justification: The main results report means with 95\% confidence-interval half-widths over random seeds, including the full main benchmark in Appendix Table~\ref{tab:main_benchmark_full_ci}; Appendix~\ref{app:uncertainty_reporting} defines the normal-approximation calculation as $1.96$ times the sample standard error of the mean.
    \item[] Guidelines:
    \begin{itemize}
        \item The answer \answerNA{} means that the paper does not include experiments.
        \item The authors should answer \answerYes{} if the results are accompanied by error bars, confidence intervals, or statistical significance tests, at least for the experiments that support the main claims of the paper.
        \item The factors of variability that the error bars are capturing should be clearly stated (for example, train/test split, initialization, random drawing of some parameter, or overall run with given experimental conditions).
        \item The method for calculating the error bars should be explained (closed form formula, call to a library function, bootstrap, etc.)
        \item The assumptions made should be given (e.g., Normally distributed errors).
        \item It should be clear whether the error bar is the standard deviation or the standard error of the mean.
        \item It is OK to report 1-sigma error bars, but one should state it. The authors should preferably report a 2-sigma error bar than state that they have a 96\% CI, if the hypothesis of Normality of errors is not verified.
        \item For asymmetric distributions, the authors should be careful not to show in tables or figures symmetric error bars that would yield results that are out of range (e.g., negative error rates).
        \item If error bars are reported in tables or plots, the authors should explain in the text how they were calculated and reference the corresponding figures or tables in the text.
    \end{itemize}

\item {\bf Experiments compute resources}
    \item[] Question: For each experiment, does the paper provide sufficient information on the computer resources (type of compute workers, memory, time of execution) needed to reproduce the experiments?
    \item[] Answer: \answerYes{}.
    \item[] Justification: The Implementation Details state that experiments were run with PyTorch on a single NVIDIA RTX PRO 6000 GPU, and Appendices~\ref{app:downstream_resource_usage} and~\ref{app:e2e_pipeline_cost} report downstream runtime/memory and end-to-end DOGMA pipeline wall-clock, GPU, RAM, and artifact-size measurements.
    \item[] Guidelines:
    \begin{itemize}
        \item The answer \answerNA{} means that the paper does not include experiments.
        \item The paper should indicate the type of compute workers CPU or GPU, internal cluster, or cloud provider, including relevant memory and storage.
        \item The paper should provide the amount of compute required for each of the individual experimental runs as well as estimate the total compute. 
        \item The paper should disclose whether the full research project required more compute than the experiments reported in the paper (e.g., preliminary or failed experiments that didn't make it into the paper). 
    \end{itemize}
    
\item {\bf Code of ethics}
    \item[] Question: Does the research conducted in the paper conform, in every respect, with the NeurIPS Code of Ethics \url{https://neurips.cc/public/EthicsGuidelines}?
    \item[] Answer: \answerYes{}.
    \item[] Justification: The work uses public single-cell transcriptomics resources and ontology databases, does not collect new human-subject data, and is evaluated as a methodological contribution for single-cell analysis.
    \item[] Guidelines:
    \begin{itemize}
        \item The answer \answerNA{} means that the authors have not reviewed the NeurIPS Code of Ethics.
        \item If the authors answer \answerNo, they should explain the special circumstances that require a deviation from the Code of Ethics.
        \item The authors should make sure to preserve anonymity (e.g., if there is a special consideration due to laws or regulations in their jurisdiction).
    \end{itemize}

\item {\bf Broader impacts}
    \item[] Question: Does the paper discuss both potential positive societal impacts and negative societal impacts of the work performed?
    \item[] Answer: \answerYes{}.
    \item[] Justification: This work aims to improve single-cell transcriptomic analysis by constructing biologically informed cell graphs from expression data and external knowledge resources. Potential positive impacts include accelerating biomedical research, improving reuse of reference atlases, reducing downstream computational cost, and helping laboratories with limited resources perform cell-type annotation, cross-condition comparison, and hypothesis generation. The method is not intended for direct clinical decision-making. Its main risks come from inherited biases or incompleteness in reference atlases, cell-type annotations, Cell Ontology, Gene Ontology, and phylogenetic or organ-level priors. Underrepresented tissues, disease states, ancestries, species, or rare cell populations may therefore receive less reliable predictions. We view DOGMA as a research tool for exploratory analysis. Responsible use should include dataset provenance checks, validation on independent cohorts, biological expert review, and audits across underrepresented cell types and populations before drawing biomedical conclusions.
    \item[] Guidelines:
    \begin{itemize}
        \item The answer \answerNA{} means that there is no societal impact of the work performed.
        \item If the authors answer \answerNA{} or \answerNo, they should explain why their work has no societal impact or why the paper does not address societal impact.
        \item Examples of negative societal impacts include potential malicious or unintended uses (e.g., disinformation, generating fake profiles, surveillance), fairness considerations (e.g., deployment of technologies that could make decisions that unfairly impact specific groups), privacy considerations, and security considerations.
        \item The conference expects that many papers will be foundational research and not tied to particular applications, let alone deployments. However, if there is a direct path to any negative applications, the authors should point it out. For example, it is legitimate to point out that an improvement in the quality of generative models could be used to generate Deepfakes for disinformation. On the other hand, it is not needed to point out that a generic algorithm for optimizing neural networks could enable people to train models that generate Deepfakes faster.
        \item The authors should consider possible harms that could arise when the technology is being used as intended and functioning correctly, harms that could arise when the technology is being used as intended but gives incorrect results, and harms following from (intentional or unintentional) misuse of the technology.
        \item If there are negative societal impacts, the authors could also discuss possible mitigation strategies (e.g., gated release of models, providing defenses in addition to attacks, mechanisms for monitoring misuse, mechanisms to monitor how a system learns from feedback over time, improving the efficiency and accessibility of ML).
    \end{itemize}
    
\item {\bf Safeguards}
    \item[] Question: Does the paper describe safeguards that have been put in place for responsible release of data or models that have a high risk for misuse (e.g., pre-trained language models, image generators, or scraped datasets)?
    \item[] Answer: \answerNA{}.
    \item[] Justification: The paper does not release a high-risk model such as a language or image generator and does not introduce scraped Internet datasets; the analyzed data are derived from public single-cell resources.
    \item[] Guidelines:
    \begin{itemize}
        \item The answer \answerNA{} means that the paper poses no such risks.
        \item Released models that have a high risk for misuse or dual-use should be released with necessary safeguards to allow for controlled use of the model, for example by requiring that users adhere to usage guidelines or restrictions to access the model or implementing safety filters. 
        \item Datasets that have been scraped from the Internet could pose safety risks. The authors should describe how they avoided releasing unsafe images.
        \item We recognize that providing effective safeguards is challenging, and many papers do not require this, but we encourage authors to take this into account and make a best faith effort.
    \end{itemize}

\item {\bf Licenses for existing assets}
    \item[] Question: Are the creators or original owners of assets (e.g., code, data, models), used in the paper, properly credited and are the license and terms of use explicitly mentioned and properly respected?
    \item[] Answer: \answerYes{}.
    \item[] Justification: Appendix~\ref{app:data_asset_licenses} lists the source studies, data portals, ontology and annotation resources, baseline repositories, license or terms-of-use information, and use in this paper. The supplementary release records dataset-specific access information and does not redistribute upstream baseline source code when no explicit redistribution license was identified.
    \item[] Guidelines:
    \begin{itemize}
        \item The answer \answerNA{} means that the paper does not use existing assets.
        \item The authors should cite the original paper that produced the code package or dataset.
        \item The authors should state which version of the asset is used and, if possible, include a URL.
        \item The name of the license (e.g., CC-BY 4.0) should be included for each asset.
        \item For scraped data from a particular source (e.g., website), the copyright and terms of service of that source should be provided.
        \item If assets are released, the license, copyright information, and terms of use in the package should be provided. For popular datasets, \url{paperswithcode.com/datasets} has curated licenses for some datasets. Their licensing guide can help determine the license of a dataset.
        \item For existing datasets that are re-packaged, both the original license and the license of the derived asset (if it has changed) should be provided.
        \item If this information is not available online, the authors are encouraged to reach out to the asset's creators.
    \end{itemize}

\item {\bf New assets}
    \item[] Question: Are new assets introduced in the paper well documented and is the documentation provided alongside the assets?
    \item[] Answer: \answerYes{}.
    \item[] Justification: The curated graph benchmarks are documented in Appendix~\ref{app:datasets}, and Appendix~\ref{app:supplementary_release} describes the anonymized supplementary ZIP containing code and reproduction configuration.
    \item[] Guidelines:
    \begin{itemize}
        \item The answer \answerNA{} means that the paper does not release new assets.
        \item Researchers should communicate the details of the dataset\slash code\slash model as part of their submissions via structured templates. This includes details about training, license, limitations, etc. 
        \item The paper should discuss whether and how consent was obtained from people whose asset is used.
        \item At submission time, remember to anonymize your assets (if applicable). You can either create an anonymized URL or include an anonymized zip file.
    \end{itemize}

\item {\bf Crowdsourcing and research with human subjects}
    \item[] Question: For crowdsourcing experiments and research with human subjects, does the paper include the full text of instructions given to participants and screenshots, if applicable, as well as details about compensation (if any)? 
    \item[] Answer: \answerNA{}.
    \item[] Justification: The work does not involve crowdsourcing, participant studies, or newly collected human-subject experiments.
    \item[] Guidelines:
    \begin{itemize}
        \item The answer \answerNA{} means that the paper does not involve crowdsourcing nor research with human subjects.
        \item Including this information in the supplemental material is fine, but if the main contribution of the paper involves human subjects, then as much detail as possible should be included in the main paper. 
        \item According to the NeurIPS Code of Ethics, workers involved in data collection, curation, or other labor should be paid at least the minimum wage in the country of the data collector. 
    \end{itemize}

\item {\bf Institutional review board (IRB) approvals or equivalent for research with human subjects}
    \item[] Question: Does the paper describe potential risks incurred by study participants, whether such risks were disclosed to the subjects, and whether Institutional Review Board (IRB) approvals (or an equivalent approval/review based on the requirements of your country or institution) were obtained?
    \item[] Answer: \answerNA{}.
    \item[] Justification: The paper does not conduct new human-subject research or crowdsourcing experiments; it analyzes public single-cell transcriptomics resources.
    \item[] Guidelines:
    \begin{itemize}
        \item The answer \answerNA{} means that the paper does not involve crowdsourcing nor research with human subjects.
        \item Depending on the country in which research is conducted, IRB approval (or equivalent) may be required for any human subjects research. If you obtained IRB approval, you should clearly state this in the paper. 
        \item We recognize that the procedures for this may vary significantly between institutions and locations, and we expect authors to adhere to the NeurIPS Code of Ethics and the guidelines for their institution. 
        \item For initial submissions, do not include any information that would break anonymity (if applicable), such as the institution conducting the review.
    \end{itemize}

\item {\bf Declaration of LLM usage}
    \item[] Question: Does the paper describe the usage of LLMs if it is an important, original, or non-standard component of the core methods in this research? Note that if the LLM is used only for writing, editing, or formatting purposes and does \emph{not} impact the core methodology, scientific rigor, or originality of the research, declaration is not required.
    %this research? 
    \item[] Answer: \answerNA{}.
    \item[] Justification: The core methodology does not use LLMs as an important, original, or non-standard component; any ordinary writing or formatting assistance, if used, would not affect the scientific method or results.
    \item[] Guidelines:
    \begin{itemize}
        \item The answer \answerNA{} means that the core method development in this research does not involve LLMs as any important, original, or non-standard components.
        \item Please refer to our LLM policy in the NeurIPS handbook for what should or should not be described.
    \end{itemize}

\end{enumerate}